# *ToxTrac:* a fast and robust software for tracking organisms

Alvaro Rodriquez[1], Hanqing Zhang[1], Jonatan Klaminder[2], Tomas Brodin[2], Patrik L. Andersson[3], Magnus Andersson[1]

[1]Department of Physics, Umeå University, 901 87 Umeå, Sweden, [2]Department of Ecology and Environmental Science, Umeå University, 901 87 Umeå, Sweden, [3]Department of Chemistry, Umeå University, 901 87 Umeå, Sweden

## Summary

1. Behavioral analysis based on video recording is becoming increasingly popular within research fields such as; ecology, medicine, ecotoxicology, and toxicology. However, the programs available to analyze the data, which are; free of cost, user-friendly, versatile, robust, fast and provide reliable statistics for different organisms (invertebrates, vertebrates and mammals) are significantly limited.

2. We present an automated open-source executable software (*ToxTrac*) for image-based tracking that can simultaneously handle several organisms monitored in a laboratory environment. We compare the performance of *ToxTrac* with current accessible programs on the web.

3. The main advantages of *ToxTrac* are: i) no specific knowledge of the geometry of the tracked bodies is needed; ii) processing speed, *ToxTrac* can operate at a rate >25 frames per second in HD videos using modern desktop computers; iii) simultaneous tracking of multiple organisms in multiple arenas; iv) integrated distortion correction and camera calibration; v) robust against false positives; vi) preservation of individual identification if crossing occurs; vii) useful statistics and heat maps in real scale are exported in: image, text and excel formats.

4. *ToxTrac* can be used for high speed tracking of insects, fish, rodents or other species, and provides useful locomotor information. We suggest using *ToxTrac* for future studies of animal behavior independent of research area. Download *ToxTrac* here: https://toxtrac.sourceforge.io

# Introduction

Perturbation in organism behavior is frequently used as a sensitive marker for stress related to social or environmental factors within various research fields of ecology and animal behavior. Behavior traits such as activity, sociality, aggression, exploration and boldness are becoming central measure within animal personality research and studies focusing on impacts of environmental changed driven evolution (Cote *et al.* 2010). Moreover, behavioral assays are becoming increasingly used within ecotoxicology (Brodin *et al.* 2014) and for medical perspective (Antunes & Biala 2012). For example, it has been argued that behavioral assays are the most sensitive tool when detecting effects of hazardous substances (Melvin & Wilson 2013), and such studies have shown behavioral effects from brominated flame-retardants (Viberg *et al.* 2013) or to bisphenol-A (Negishi *et al.* 2014). Indeed, multiple research fields have a great need to accurately measure organism behavior and many of the unexpected biological effects from xenobiotics in the environment could potentially have been detected before their introduction on the market (and the environment) if behavioral assays had been used in routine chemical risk assessment procedures. Laboratory assays using visual tracking involving critical behavioral patterns are evolving; however, current methodologies generate huge amounts of data and are often laborious and time-consuming. Available software to analyze this kind of data are costly, slow, often requires programming skills, or are developed as plug-in functions to proprietary programs.

Important features of a tracking software for monitoring animal behavior are algorithms that accurately can detect the position of the organisms and provide a reliable analysis for long time series. Also, it should allow handling of new experimental setups. In addition, the software should be: adjustable to new experimental setups; robust over time and computationally efficient, preferably enabling data processing of several experiments simultaneously. A rather efficient tracking algorithm is the Kalman filter, which also is the most common algorithm used for tracking (Welch & Bishop 2006; Rodriguez *et al.* 2014). This recursive filter uses the animals previous position to estimate the next, thus reducing the search range and the risk of identification foreign objects as the tracked animals. The main advantages of the Kalman filter are that it allows tracking of multiple objects and it does not require any knowledge of the animal shape. However, the Kalman filter always follows the closest known object to the predicted position, which means that it is very sensitive to false-positives and false-negatives. Additionally, the Kalman filter is unreliable in the presence of occlusions since it does not preserve the identity of multiple animals after occlusions.

Different approaches have been developed to address the problem of tracking multiple animals. Using several cameras can improve tracking by using data from different perspectives to solve occlusions (Maaswinkel *et al.* 2013). However, this adds complexity to the experimental setup and increases the amount of data. Some techniques tag individuals with visual markers to provide a distinct ID to each animal (Crall *et al.* 2015); other techniques use specific features of the animals such as the symmetry axis (Fontaine *et al.* 2008) or the shape of the head (Qian *et al.* 2014, 2016). Also model based approaches that use the geometry of the body are suggested (Wang *et al.* 2016). The work by Perez-Escudero et al. is a notable attempt to keep the identities correct overtime (Pérez-Escudero *et al.* 2014). They extract characteristic fingerprints from each animal that are matched with the trajectories in the video. This approach is, however, very computationally heavy and as such unsuitable for real-time applications; online processing and for bulk processing of huge data sets. In addition: it relies on a non-general separation stage; it is not scalable with regards of the numbers of animals, the video length or the video resolution; and finally, it requires a minimum size of the individuals and a huge number of samples for each individual to work (3000 samples are recommended).

Thus, there is a need for a program that is open-source, easy to use and flexible, robust overtime, can handle multiple species, and analyze data fast. In this work, we present *ToxTrac*, a free program for tracking animals in multiple arenas. *ToxTrac* can also track multiple animals reliably handling occlusions and preserving the identity of the individuals.

## Performance and results of the program

To evaluate the performance of *ToxTrac*, we measured the ability to detect and track individuals using several species and we tested the preservation of identity of multiple individuals in one labeled dataset. To compare the performance of *ToxTrac* with other software's, we downloaded the free tracking tools: *MouseMove* (Samson *et al.* 2015), an open source program for semi-automated analysis of movement in rodents; *Idtracker* (Pérez-Escudero *et al.* 2014), focused in tracking multiple organisms and keeping the correct identities; *BioTrack* (Feldman *et al.* 2012; Hrolenok *et al.* 2012)*,* which uses shape models to track rigid organisms; *EthoWatcher* (Junior *et al.* 2012)*,* developed for the extraction of kinematic variables of laboratory animals; *Ctrax* (Branson *et al.* 2009, 2015), a tool developed for walking flies; and finally *SwisTrack* (Correll *et al.* 2006; Mario *et al.* 2014). The main functions of these software tools are presented in Table S1.

Based on this we conclude that *BioTrack* can only be used on a very limited and specific range of species since it assumes a rigid displacement, while many animals, such as fish, are characterized by having deformable bodies that cannot be matched by a rigid descriptor. *Ctrax* is also not suitable for tracking some types of animals, since it uses an ellipse fitting step which tends to separate certain shapes into multiple individuals. *EthoWatcher* and *MouseMove* cannot analyze multiple arenas or multiple animals and in addition, *EthoWatcher* it restricted to a video resolution of 320x240 pixels, whereas *MouseMove* is restricted to a resolution of 640x480 pixels. We also find that *SwisTrack* is unreliable since it is unstable and fails to track more than one animal in videos with separated arenas, or when the animals stay stationary. Therefore, *Idtracker* is the only reliable open-source tracking software currently available. We therefore compared, when possible, the results obtained using *ToxTrac* with that of *Idtracker.*

We first evaluated the detection rate of *ToxTrac* and *Idtracker*, defined as the mean percentage of frames where each individual was correctly detected by the system. We used 8 video datasets from 7 organisms sampled using different illumination conditions: 4 ants (*Formica rufa*) placed in an arena with direct illumination (Figure 1a), a cockroach (*Blaptica dubia*) in an arena with direct illumination (Figure 1b), 4 Trinidadian guppy (*Poecilia reticulata*) in 4 arenas with backlight illumination (Figure 1c), 2 Trinidadian guppy in an arena with direct illumination (Figure 1d), a dataset obtained from (Samson *et al.* 2015) with one C57BL/6 mice (*Mus musculus*) in an arena with diffuse illumination (Figure 1e), 3 Atlantic salmon (*Salmo salar*) in 3 arenas with backlight illumination (Figure 1f). 4 tadpoles (*Rana temporaria*) in 4 arenas using direct illumination (Figure 1g), and a dataset obtained from (Pérez-Escudero *et al.* 2014) with 5 zebrafish (*Danio rerio*) in a single arena with diffuse illumination (Figure 1h). Four arenas with tadpoles are shown in Figure S1 and the detailed characteristics of the datasets are shown in Table S2.

*ToxTrac* and *IdTracker* showed an average detection rate of 99.2% and 95.9% respectively in the non-occluded experiments. As expected, when the animals are occluded, that is, when multiple individuals are in the same spot and they are not visually separable, the detection rate drops accordingly. In this case *IdTracker* obtains better results. This is due to a resegmentation stage that *IdTracker* uses to improve its performance in experiments with multiple animals with fish-like shape.

The tracking times were computed for *ToxTrac* and *IdTracker* in the same computer, both algorithms used a 500 reference frames for storing feature information and other parameters were set as similar as possible. We measured the processing times of *ToxTrac* with and without using the fragment identification post processing step. The tracking times with *ToxTrac* are in all cases significantly lower.

It should be also noted that *IdTracker* was not able to handle the cockroach experiment with its original resolution, and presents a very poor performance in high-resolution videos. The complete results of this experiment are presented in Table 1.

We also evaluated the fragment identification algorithm. We used the zebrafish dataset (Pérez-Escudero *et al.* 2014), with 5 zebrafish showing several crossings with each other. We manually labeled the tracks obtained from the video, analyzing each crossing to determine which tracks that corresponded to which individual, and compared this data with the *ToxTrac* results. The detections metrics were obtained counting the number of the detections of the correctly and incorrectly assigned tracks. Figure 2 shows some examples of how the identification algorithm solves occlusions, and Table 2 shows the complete results of this experiment.

*ToxTrac* was able to preserve the identity of the tracked animals in 99.6% of the cases when tracks were longer than 50 frames. For shorter tracks, the identification algorithm is less reliable, but still shows a very good accuracy, 92.2%. Though a comparison under the same conditions is not possible with *IdTracker,* due to the differences of the workflow of both algorithms, the reported results of *IdTracker* and other state-of-the-art-techniques using fragment linking algorithms (Pérez-Escudero *et al.* 2014; Qian *et al.* 2014) are similar. However, *ToxTrac* has the advantage of not requiring a detection algorithm specific for a particular animal shape and of being significantly faster.

Explanation and a flow diagram of the sub-routines used in the tracking algorithms are found in supplementary materials 1 and Figure S2. A complete user guide for the program is provided in supplementary materials 2.

# Discussion

Animal behavior is central in ecological and evolutionary research and is receiving increased attention in many other fields such as ecotoxicology, medicine, neurology and toxicology. Indeed, *ToxTrac* provides several important criteria for becoming a useful tool for these fields being increasingly dependent on animal tracking. The free *ToxTrac* software demands no specific knowledge of the geometry of the tracked objects, can conduct real-time processing of behavioral data, can simultaneously track multiple organisms in multiple arenas, integrate distortion correction and real scale calibration measurements and functions robustly against false positives and preserves the identity of individuals if crossing or occlusion occurs.

Some brief examples of endpoints of ecological relevance that *ToxTrac* can measure are outlined below. *ToxTrac* is able to measure behavioral parameters necessary for calculating metabolic rates such as locomotor activity (average speed, acceleration and distance traveled per time unit). The program can also be used to measure the time an organism spends near aquaria or terrarium walls which is a common measure of anxiety within ecotoxicology (Maximino *et al.* 2010). In *ToxTrac* the user can adjust the distance of interest to any set object. The flexibility in regions of interest makes the program directly applicable in tests such as scototaxis (dark/light preference), commonly used as a measure of boldness or anxiety but also validated for assessing the antianxiety effects of pharmacological agents (Maximino *et al.* 2010). The visual plots allows assessment of trajectories that can be used as a measure of how efficient individuals are in exploring novel areas and such data, combined with the generated information of boldness and activity, are necessary when trying to understand how behavioral traits are linked to individual performance or vulnerability to change (e.g. pollution) (Conrad *et al.* 2011). Here, the output of all the measures in a widespread format (excel spread-sheet) makes it easy for the user to export data into other format necessary for further analysis. The near universal application of the software is also evident from our results showing that the program allows tracking of organisms of very different morphologies and movement patterns (i.e. ants, cockroaches, Atlantic salmon, tadpoles of the common frog, zebrafish and the Trinidadian guppy). In our assays, *ToxTrac* was capable detecting the animals in 99.9% of the frames, which is considered highly acceptable for the common application of behavioral assays. Although *ToxTrac* originally was developed to assay one animal per arena it is still capable of handling occlusions and multiple animals preserving the identity of the animals using a fragment linking algorithm. Our analysis shows that *ToxTrac* is at least as accurate as other programs used for tracking multiple individuals simultaneously but it process the data significantly faster.

From a practical perspective, *ToxTrac* handles image distortion by using a simple calibration process, which facilitates tracking in environments where, for example, novel objects are co-occurring in images. It also provides bulk processing capabilities, so it is possible to analyze experiments with hundreds of videos in a single step, which greatly facilitate handling large data sets that typically accompany behavioral assays within ecology.

## Acknowledgement

This work was supported by the Swedish Research Council (2013-5379), the Kempe foundation and from the ÅForsk foundation.

# Tables

Table 1: Tracking performance of *IdTracker* and *ToxTrac* for eight datasets from seven different organisms.

| Experiment | Detection Rate | | Tracking Time | | |
|:---:|:---:|:---:|:---:|:---:|:---:|
| | *IdTracker* | *ToxTrac* | *IdTracker* | *ToxTrac* *(With Id. Preservation)* | *ToxTrac* *(Without Id. Preservation)* |
| *Ant* | 98.25% | 99.99% | 00:17:20 | 00:04:12 | 00:04:06 |
| *Cockroach* | 99.99% | 99.99% | 01:00:30[1] | 00:12:20 | 00:10:54 |
| *Guppy 1* | 96.94% | 97.65% | 00:53:38 | 00:10:42 | 00:09:23 |
| *Guppy 2* | 99.11% | 94.52% | 00:02:44 | 00:00:06 | 00:00:04 |
| *Mice* | 99.99% | 99.99% | 00:02:13 | 00:00:15 | 00:00:14 |
| *Salmon* | 90.55% | 99.52% | 01:14:10 | 00:18:33 | 00:16:43 |
| *Tadpole* | 99.99% | 99.98% | 00:03:16 | 00:00:18 | 00:00:16 |
| *Zebrafish* | 93.23% | 87.36% | 00:42:32 | 00:09:53 | 00:02:31 |

[1]Dataset analyzed using a resampled image size (half the original size) to reduce the tracking time.

Table 2: Identity preservation performance of *ToxTrac*.

|  | **Total** | **Correct Id. Rate** | **Incorrect Id. Rate** | **Not Id. Rate** |
|---|---|---|---|---|
| *Long Tracks* | 282 | 99.65% | 0.35% | 0.00% |
| *Short Tracks* | 115 | 92.17% | 5.22% | 2.61% |
| *Detections* | 65,518 | 99.49% | 0.37% | 0.14% |

# Figures

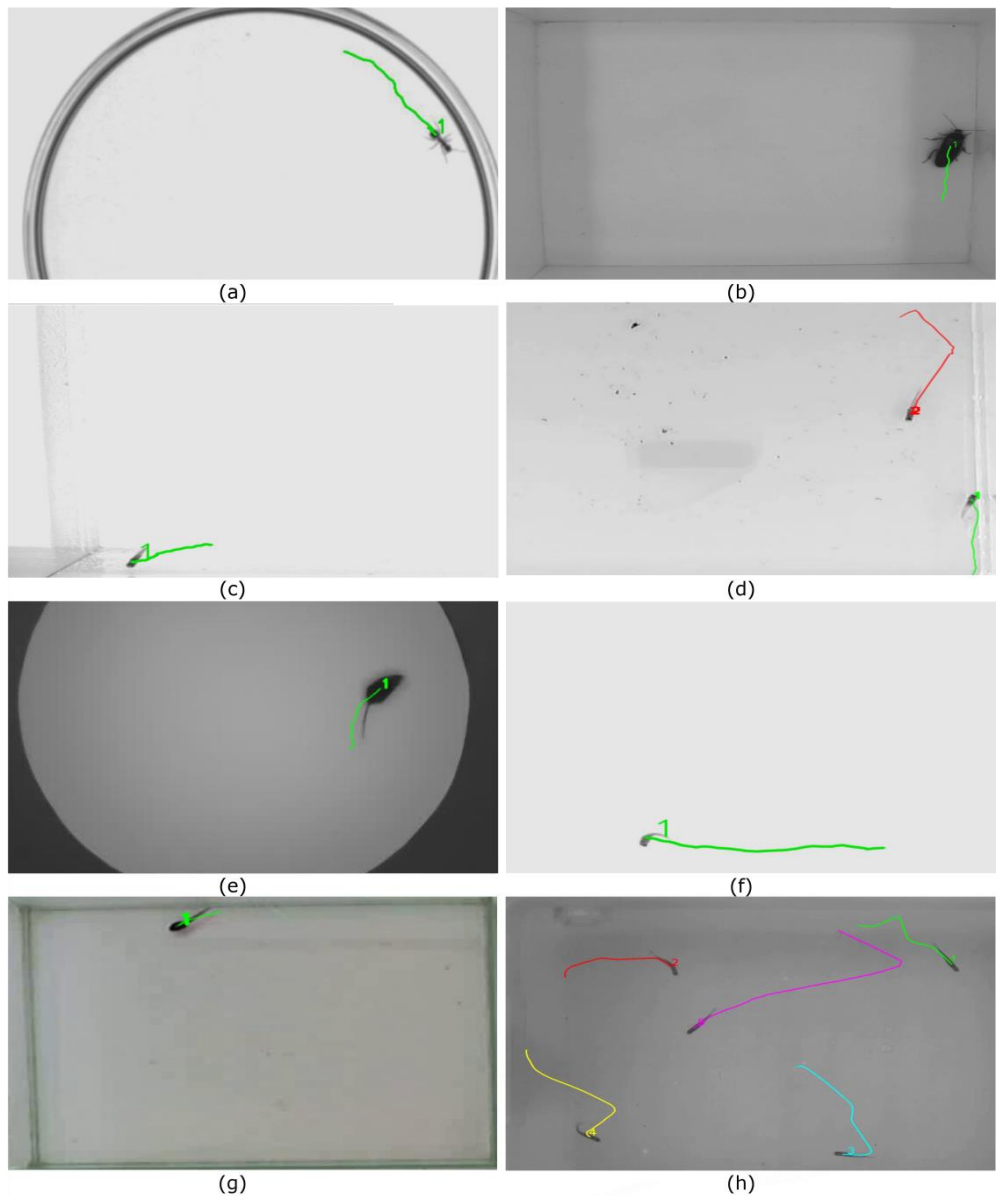

(a)

(b)

(c)

(d)

(e)

(f)

(g)

(h)

Figure 1. *ToxTrac* tracking examples using eight different organisms illuminated using either backlight or direct light. (a) ant, (b) cockroach, (c) guppy 1, (d) guppy 2, (e) mouse, (f) salmon, (g) tadpole, (h) zebrafish.

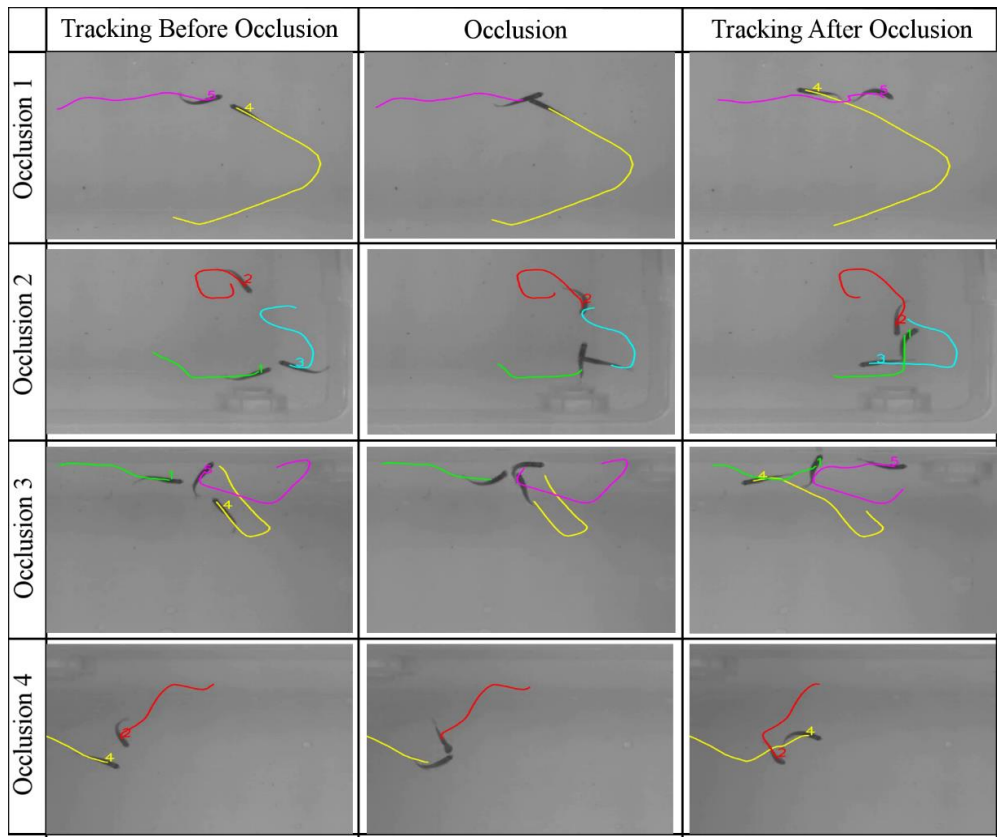

Figure 2. Example of occlusions handled by *ToxTrac* from the zebrafish dataset (Pérez-Escudero *et al.*

2014).

# *ToxTrac:* a fast and robust software for tracking organisms

# User guide for version 2.61
(Submitted to the Journal Methods in Ecology and Evolution)

Alvaro Rodriquez[1], Hanqing Zhang[1], Jonatan Klaminder[2], Tomas Brodin[2], Patrik Andersson[3], Magnus Andersson[1]

[1]Department of Physics, Umeå University, 901 87 Umeå, Sweden, [2]Department of Ecology and Environmental Science, Umeå University, 901 87 Umeå, Sweden, [3]Department of Chemistry, Umeå University, 901 87 Umeå, Sweden

# Contents

# Starting Guide

## 1. Requirements

*ToxTrac* has been developed for Windows in C++ using Visual Studio 2015 with the add-on Qt5Package. In addition we have used OpenCv3.0, an open-source computer vision library available at http://opencv.org, Qt5.6.0 open-source, a library for building interfaces available at https://www.qt.io, and LibXL, a library for writing and reading excel files available at http://www.libxl.com.

*ToxTrac* requires Windows 7 or later, and is created for 64-bit hardware. We recommend a minimum of 8 GB of RAM memory and enough hard drive free space to handle all recorded video files. A 2.0+ GHz Quad core or higher is recommended for proper performance.

## 2. Installing and running the software

The *ToxTrac* project, containing the latest software version, the documentation of the program and other resources is hosted at https://toxtrac.sourceforge.io.

To install *ToxTrac*, it is only necessary to execute the .exe Windows installer file provided.

Visual Studio 2015 or Visual C++ x64 2015 Redistributable Packages (or newer) are required to run *ToxTrac*. The installer will automatically download and install this component if necessary.

Once Visual Studio or the Visual C++ Redistributable Package is installed, follow the instructions of the install process. *ToxTrac* and all necessary components will be installed in the selected folder and a shortcut in the Desktop and the start menu will be created. Now *ToxTrac* is ready.

An updated codec pack is recommended to properly handle video files with *ToxTrac*. We have used K-Lite Codec Pack, available at https://www.codecguide.com.

It is also recommendable to install any spreadsheet package compatible with .xls (Microsoft Excel files) to view the statistical file generated by *ToxTrac*.

To inspect and edit video files we recommend VirtualDub, available at http://virtualdub.org.

To inspect and edit image files, we recommend GIMP, available at https://www.gimp.org.

## 3. Known Issues

- A minimum screen resolution of 1280x960 is required to display properly the interface windows.
- *ToxTrac* is aimed to work with 96 dots per inch screens (the Microsoft Windows operating system default display). Changing the dpi, for example to increase the font size in the windows accessibility configuration, will not increase the text size. Instead, *ToxTrac* will try to rescale all text elements to avoid clipping. This behavior may result in difficulties to visualize the interface for some screens, especially in tablets or some portable devices.
- *ToxTrac* does not recognize non-standard folder of file names. Though spaces are permitted, the program will now work properly if: The used file names, the project, the calibration or the video folders contain non English characters.
- Using the fragment identification algorithm may require a lot of RAM memory. If the used memory exceeds the system capacity, *ToxTrac* may crash. The fragment identification algorithm can be configured to avoid this issue, (for example using a TCM depth of 8 pixels, reducing the TCM radius or the maximum number of samples per fragment). Additionally, changing the maximum number of fragments will force the program to free the memory when the number of trajectory fragments in memory reaches a certain limit. The identification algorithm can be partially or totally deactivated to reduce computational time and prevent any memory problem.

- If the project location is in a folder without write permission, *ToxTrac* will not be able to save the project data and will show a warning video.

## 4. Recording

### Video Files

*ToxTrac* supports a wide variety of .avi video files with any resolution and framerate, including MPEG-4 and x264 compressed .avi video files. However, the video should of course have the highest possible quality.

If the video is cut in several files, all the pieces should be placed in the same folder and file names should end in a correlative number according to the file order (example: video01.avi, video02.avi, …).

Multiple video sequences (each sequence composed by multiple video files) can be analyzed at the same time (if experimental conditions are not changed).

Video resolution should be high enough so that the animal size is at least 50 pixels, and framerate should be high enough so the animal area in consecutive frames overlaps. We find that 25fps is enough for most experiments, but with fast moving animals, and especially in multiple animals experiment, a higher framerate may be advisable.

*ToxTrac* will automatically convert the video image format to a grayscale image, so color information is not needed.

### Tracking areas and background

*ToxTrac* will detect and track animals in rectangular pieces of the image containing the arenas where we want to observe the animals. Inside the arena, the tracking areas are defined as uniform bright regions with no particular shape, where the tracked objects can be detected. If the arenas have dark corners or edges, these should be excluded from the tracking area.

Ideally, the background color in the tracking areas should as homogeneous as possible, and brighter than the animal, with the highest possible contrast.

The presence of different background objects (dark objects appearing inside the tracking areas) is acceptable, and will be managed by the system in different ways: Static or moving objects much smaller than the animals will not affect segmentation, because they can be filtered out. Static objects of any size can be removed using the background subtraction technique (this may cause animals not to be detected until they start to move). Alternatively, the arena selection tool can be used to exclude parts of the background.

Objects which cannot be separated from the animals (moving objects, or static objects which are not excluded from the arena) will difficult or impede the tracking process. It is especially important to use a background as free of reflections or shadows as possible (see lightning) and to exclude the dark edges of the experimental setup from the background (see arena definition).

### Animals

Multiple animals can be tracked in the same arena, however, we don't recommend using more than 10-20 animals in a single experiment if it is important to keep the identity of the animals during the experiment. This value however depends on the occlusion degree observed in the video.

## 5. Experimental Setup

The experimental setup is typically formed by the arenas, the camera and the illumination elements (lights, filters, diffusers…). During an experiment, the experimental setup should be isolated from external interference and external light variations.

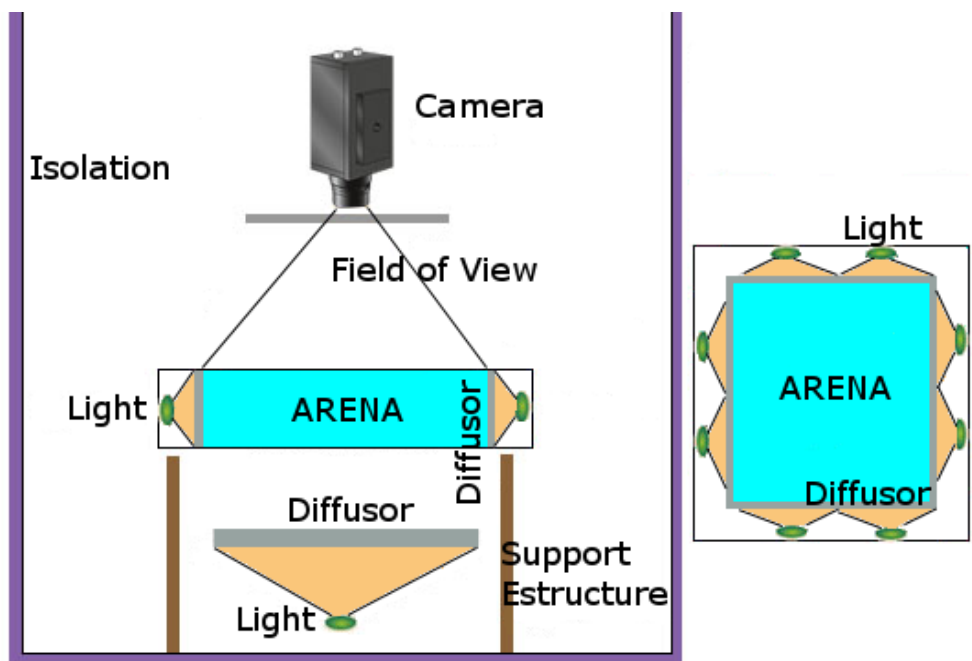

Figure 1. Arena schematic.

It is very important that the experimental setup present the same conditions and it is not moved during an experiment, between the calibration and the experiment, or between experiments, if the same calibration data is used or if they are going to be processed together.

## Arena

The walls of the arena should not cast strong shadows or reflections. This can be achieved by a studying the camera position, the lightning conditions and the arena materials. Transparent or translucid walls will not cast shadows, but may have reflections; and opaque walls will not have reflections but may cast shadows. Also wall height can be adjusted to reduce these effects, and walls can be in most cases excluded from the tracking area.

## Illumination

Without an appropriate illumination, the task of tracking is impossible. The light conditions determine the typo, position, angle and intensity of the beams incident in the arenas which will be then registered by the camera.

We recommend two types of illumination to be used in the tracking system:
-   Diffuse illumination, this light preserves the texture details and mitigates shadows.
-   Backlight illumination, this light will highlight the shapes and will not cause shadows. However, it will also hide the textures of the objects.

A scheme of how to construct these illumination types is shown in Figure 2, and Table 1 shows how each type of light interacts with different characteristics of the objects.

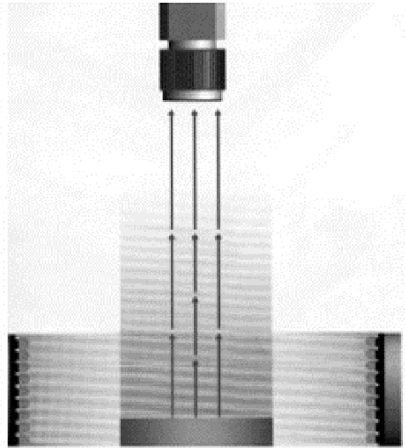 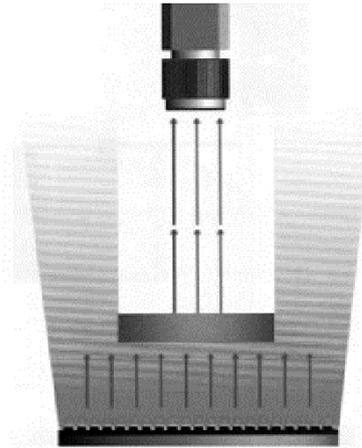

Dark field (diffuse) illumination      Backlight illumination

Figure 2. Dark filed and backlight illumination.

Table 1. Dark filed and backlight illumination characteristics.

| Feature | Backlight | Dark Field |
|---------|-----------|------------|
| Absorption (Changes in light absorption from the object) | None | Minimal effect |
| Texture (Changes un surface texture) | None | Textured surfaces brighter than polished |
| Elevation (Changes in height from surface to camera, z axis) | None | Outer edges are bright |
| Shape (Change on shape or contour along x/y axis) | Shows outside contours | Contours highlighted, flat surfaces darker than raised |
| Translucency (Changes in density-related light transmission) | Shows changes in translucency vs. opaqueness | None |

# Basic Functions

## 1. Start Screen

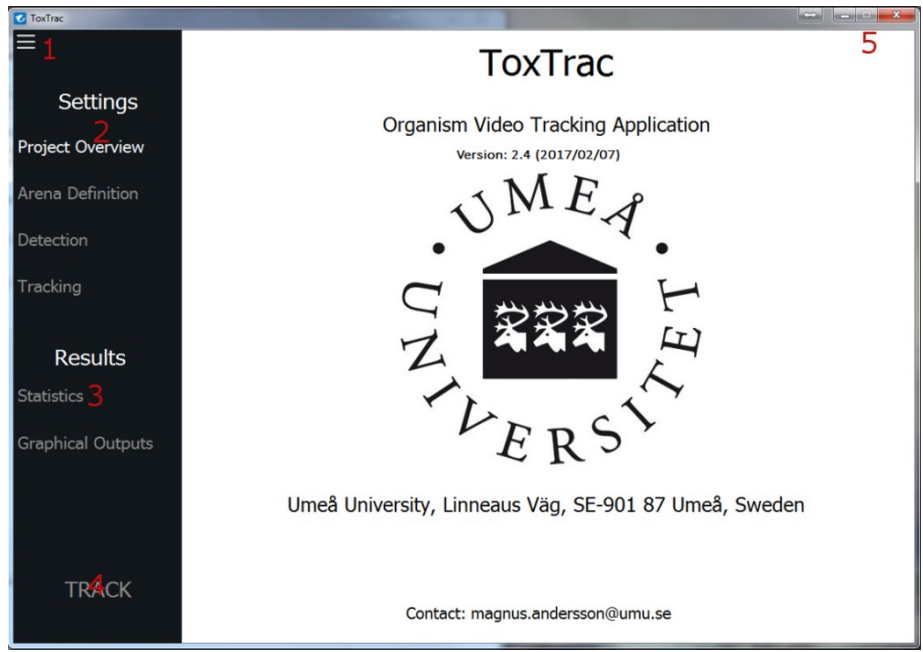

Figure 3. Start Screen.

**1:** Access to the load/save menu, allowing different options: "**New Project**" (deletes all current data, and reloads color and configuration files), "**Save Project**", "**Load Project**" or "**Merge Project**" in the current one (merges another project in the current one, this requires that both projects are recorded in the same conditions, with the same calibration data and the same arena definition parameters).

**2:** Settings menu, this allows the user to navigate through the different settings of the software. These are divided into four different panels: "**Project Overview**", "**Arena Definition**", "**Detection**", and "**Tracking**".

**3:** Results menu, (only available after analyzing the video sequences), this allow the user to see the statistical data of the video sequences and the graphical outputs.

**4:** Tracking/Stop**,** Process the videos with the current configuration. It automatically saves the project when the analysis starts and when it finishes. During the analysis, this button will allow the user to stop the current processing, in a safe point in the body tracking or in the fragment identification algorithm. If a process is stopped, all progress is lost.

**5:** Window Menus, Standard windows buttons to minimize, maximize and close *ToxTrac* main window.

## 2. Project Files

A project is a coherent tracking analysis, identified by the project name (*out.pnam*). One unique calibration and configuration will be used in one project. One project can have several video sequences and one video sequence can have several video files. All sequences are expected to be recorded in the same experimental conditions and with the same camera parameters.

When pressing the Analyze button, every sequence in the project will be analyzed, and the results will be located in a set of subfolders in the project folder. Also, all results will be joined to estimate the statistics of the entire population analyzed.

Table 2. Project files

| Project File | Example of the File | Description |
|---|---|---|
| pname.tox | C:/ProjectFolder/pname/pname_Input.txt<br>C:/ProjectFolder/pname/pname_Configuration.txt<br>C:/ProjectFolder/pname/pname_Arena.txt<br>C:/ProjectFolder/pname/pname_ArenaNames.txt<br>C:/ProjectFolder/pname/pname_Calibrator.txt<br>C:/ProjectFolder/pname/pname_Output.txt | Main project file.<br>Contains a link to Input, output and configuration files<br>Is generated by the application when saving or analyzing, and can be loaded in the application<br>Can be added to another compatible project, joining both populations. |
| pname_Input.txt | 1<br>2<br>C:/VideoFolder/Seq1_0.avi<br>C:/VideoFolder/Seq1_1.avi<br>0 2054 | Video input file.<br>Contains the information of the video sequences.<br>For each video sequence: the corresponding video files<br>A video and frame number is stored to define a reference frame where arenas will be defined. |
| pname_Configuration.txt | CALIBRATION_PARAMETERS […]<br>ARENA_DEFINITION_PARAMETERS […]<br>BACKGROUND_PARAMETERS […]<br>PREPROCESSING_PARAMETERS […]<br>DETECTION_PARAMETERS […]<br>KALMAN_FILTER_PARAMETERS […]<br>KALMAN_MULTITRACKING_PARAMETERS […]<br>DATA_ANALYSIS_PARAMETERS […]<br>MAIN_VIDEO_PARAMETERS […] | Configuration file.<br>Contains all the main parameters used in the program.<br>Some of these parameters are accessible in the interface.<br>These parameters can be modified in the text file, to modify the algorithms behavior. |
| pname_Arena.txt | 1<br>200 4 1522 1075 | Arena definition file<br>Contains the corner coordinates in pixels, of the manually drawn arenas. |
| pname_ArenaNames.txt | 1<br>Arena1 | Arena Names file<br>Contains the names of the arenas introduced by the user |
| pname_Calibrator.txt | […]<br>1     1     0<br>10     0     0<br>0     9.5     0<br>0     0     1<br>[…] | Calibration parameters file.<br>Contains the camera matrix, the rotation matrix, the displacement vectors and the distortion coefficients of the calibration.<br>This file is used to scale the primary output and also to remove the distortion from every video frame. |
| pname_Output.txt | 1<br>2<br>C:/ProjectFolder/ Seq1 / Tracking_0.txt<br>C:/ProjectFolder/ Seq1 / Tracking_1.txt | Primary output file.<br>Points to the tracking results.<br>All results will be extracted from these files by the application.<br>This output will generate a file for each arena and sequence. |

# Project Overview

## 1. Project overview screen

This screen shows the selected videos to track, and allows to add, remove, or reorder them. In addition, the user can modify basic parameters.

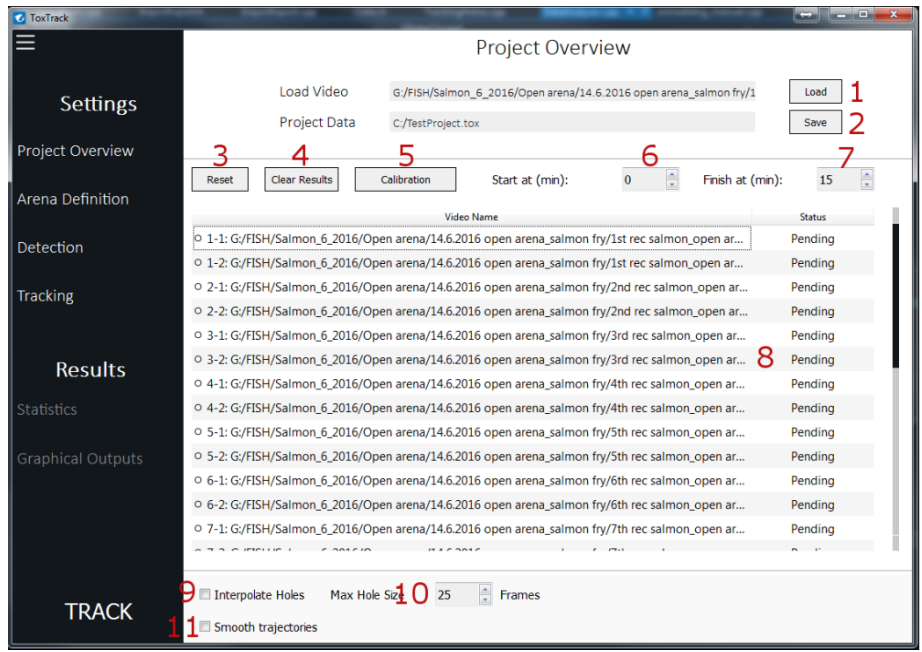

Figure 4. Project overview screen.

**1:** Load a new video sequence in the list. To add a sequence formed of multiple videos named with correlative numbers, select the first video of the sequence and all files representing video fragments of the same video will be added to the list as a video sequence. Each sequence will be analyzed as one entity.

**2:** Select project location in the hard drive, all results will be saved in this folder, be sure you have writing permissions in this folder.

**3:** Restart the project, removes all results, videos, and calibration data, keeping current configuration.

**4:** Remove all results computed for current videos.

**5:** Open calibration window.

**6:** Select the starting point in mins (for all sequences) for the analysis.

**7:** Select the ending point in mins (for all sequences) for to analysis.

**8:** List of videos, right click in one of the videos allows to move up or down, or to delete the corresponding sequence in the list. It also shows the status of the current video, and the completion rate of the different stages of the process.

**9:** Interpolate holes in the trajectory, using a linear interpolation algorithm. This parameter can be changed without redoing the analysis.

**10:** Maximum size of a trajectory hole (in frames) where interpolation will be applied. This parameter can be changed without redoing the analysis.

**11:** Use a moving average to smooth trajectories. This parameter can be changed without redoing the analysis.

## 2. Calibration algorithm

### Camera model

In practice, due to small imperfections in the lens and other factors, some distortions are integrated into the image. These distortions can be modeled using the following parametric equations[2]:

$$d(x) = x\frac{1 + k_1 r^2 + k_2 r^4 + k_3 r^6}{1 + k_4 r^2 + k_5 r^4 + k_6 r^6} + 2p_1 xy + p_2\left(r^2 + 2x^2\right) + s_1 r^2 + s_2 r^4 \qquad (1)$$

$$d(y) = y\frac{1 + k_1 r^2 + k_2 r^4 + k_3 r^6}{1 + k_4 r^2 + k_5 r^4 + k_6 r^6} + p_1\left(r^2 + 2y^2\right) + 2p_2 xy + s_3 r^2 + s_4 r^4 \qquad (2)$$

$$r^2 = x^2 + y^2 \qquad (3)$$

where $x$ and $y$ are spatial coordinates, $r$ is the distance to the lens optical center, $d(x)$ and $d(y)$ are the corresponding distorted coordinates, $k_i$ are the radial distortion coefficients, $p_i$ are the tangential distortion coefficients and $s_i$ the prism distortion coefficients.

The distortion coefficients do not depend on the scene viewed, thus they also belong to the intrinsic camera parameters. And they remain the same regardless of the captured image resolution.

### Calibration using patterns

Calibration is performed using a sequence of images of a calibration pattern. The calibration images should be recorded using the same conditions as in the intended experiment and in the same image resolution. The calibration pattern has a shape of a black and white chess-board pattern, without edge lines and with white edges. It must be printed in high resolution, and can be automatically generated in some image editing tools such as GIMP. The images should be taken in a well illuminated environment, in the plane of interest of the experiment, and using the same camera position and camera parameters as in the actual experiment. The calibration images should be placed in the same directory and with a file name ending in correlative numbers. An example of calibration pictures can be seen as follows.

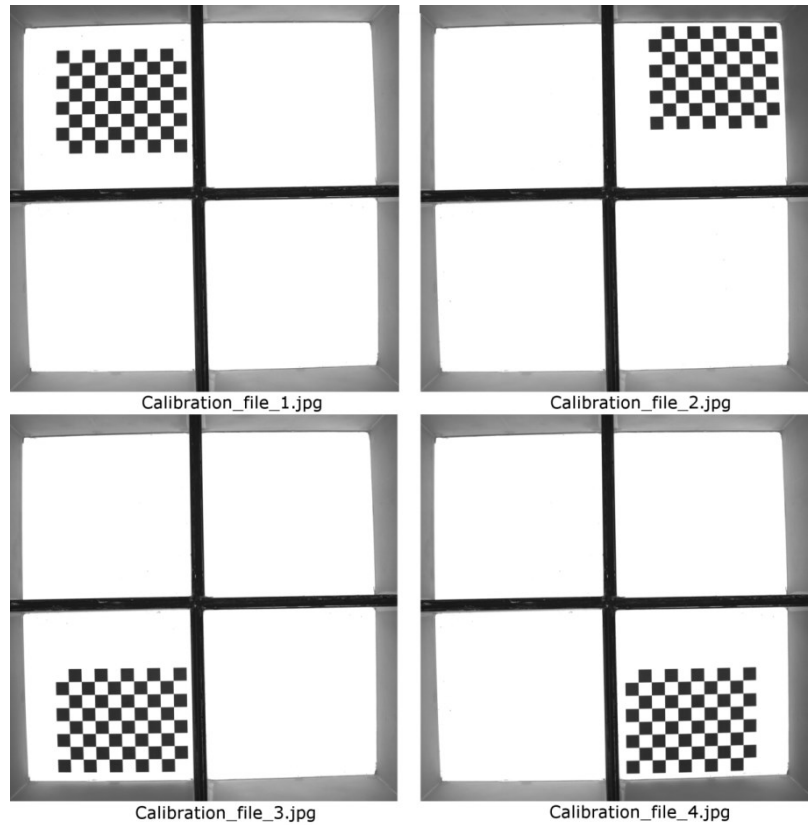

Figure 5.  Example of calibration views taken for a 4 arena experiment.

The calibration proceeds as follows:

The user first selects the first image of the sequence with *ToxTrac*.

The user selects one of the calibration images to define the camera pose.

A corner detection algorithm finds the positions of all the squares in the images. If the found positions do not correspond with the number of rows (*cal.rows*) or columns (*cal.cols*) of the pattern for a particular image, the algorithm will return an error, and that image should be eliminated from the sequence.

The algorithm will use the known dimensions of the squares of the calibration pattern (*cal.size*) and their positions in the image to estimate the calibration parameters using the global Levenberg-Marquardt optimization algorithm.

The user can select different distortion models (*cal.dist*), using different subsets of the distortion coefficients from equations 8-10. Available models are: "***Radial 3***" (dist. coeffs.: $k_1$, $k_2$, $k_3$), "***Radial 3 + Tangential 2***" (dist. coeffs.: $k_1$, $k_2$, $k_3$), "***Radial 6 + Tangential 2***" (dist. coeffs.: $k_1$, $k_2$, $k_3$, $p_1$, $p_2$, $k_4$, $k_5$, $k_6$), "***Radial 6 + Tangential 2 + Prism 4***" (dist. coeffs.: $k_1$, $k_2$, $k_3$, $p_1$, $p_2$, $k_4$, $k_5$, $k_6$, $s_1$, $s_2$, $s_3$, $s_4$).

## Manual calibration

When it is not possible to obtain views of the calibration patterns, calibration can be performed manually, introducing the parameters of a camera model.

The most straightforward way to do this is to assume that the alignment error and the distortion of the camera can be disregarded. In this case, translation vectors, distortion parameters, and the Euler pose rotation vectors should be set to 0. The calibration proceeds as follows:

-   The user measures the horizontal and vertical mm to pixel scale of the image. For example a horizontal scale of 1:10 (1 mm = 10 pixel) and a vertical scale of 1:9.5 (1 mm = 9.5 pixel).

- The scale factors are introduced in the camera matrix parameters *fx* and *fy*. The camera matrix for the example values is shown in equation 11.

$$M = \begin{bmatrix} 10 & 0 & 0 \\ 0 & 9.5 & 0 \\ 0 & 0 & 1 \end{bmatrix}, \qquad (4)$$

## 3. Calibration screen

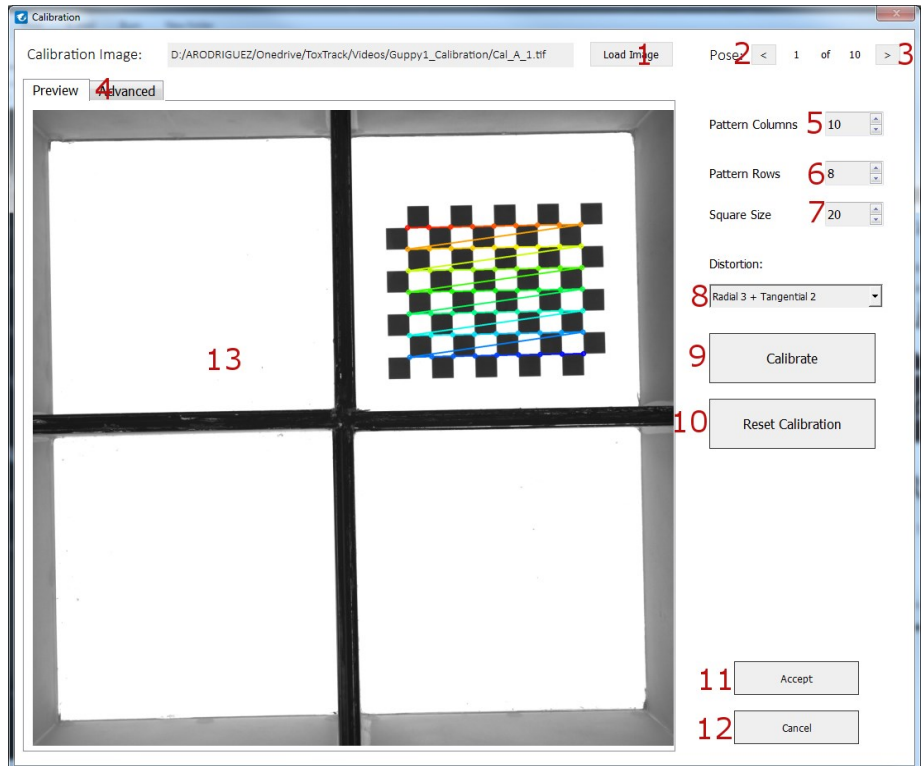

Figure 6. Calibration screen, preview.

**1:** Select a calibration sequence. The user must select the first the first image of the sequence. The calibration images must have a name ending with correlative numbers and should be placed in the same directory

**2:** Select previous image for pose estimation.

**3:** Select next image for pose estimation.

**4:** Show/hide the advanced calibration parameters.

**5:** Number of columns of the calibration pattern.

**6:** Number of rows of the calibration pattern.

**7:** Dimensions in mm of the squares of the calibration pattern.

**8:** Distortion model.

**9:** Try to estimate calibration parameters using the selected calibration file, and the current parameters.

**10:** Reset the calibration parameters.

**11:** Accept calibration.

**12:** Cancel, closes the window, no changes will be made in the calibration data.

**13:** View of the current image of the calibration sequence. If calibration if successful, it will show the detected features of the pattern.

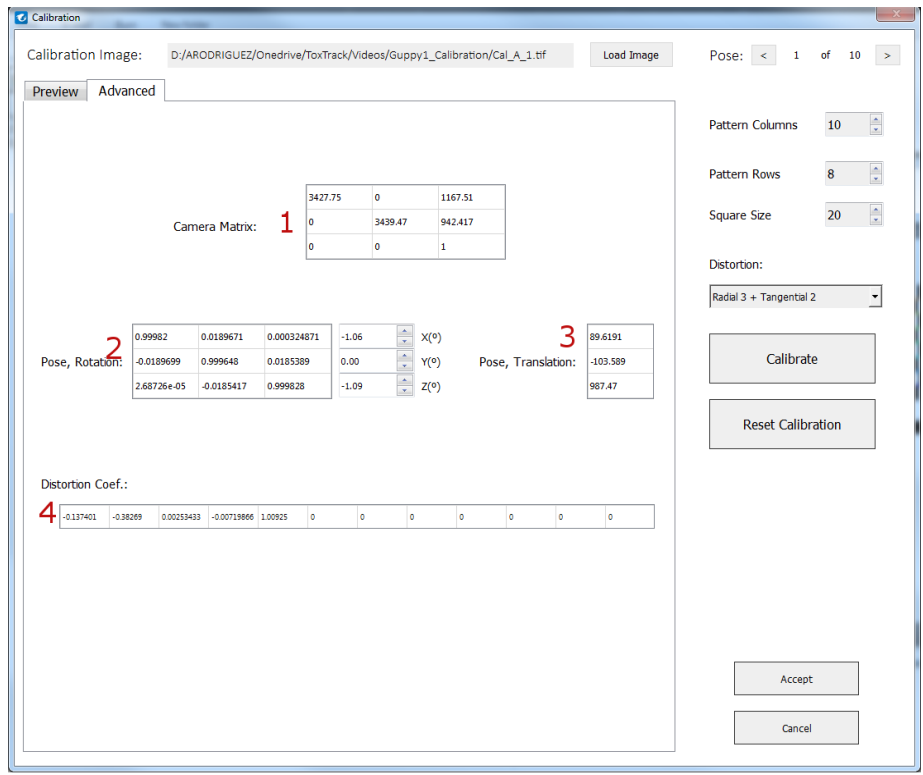

Figure 7. Calibration screen, advanced.

**1:** Camera matrix, the pin-hole projection parameters of the camera.

**2:** The rotation matrix, is used to determine the orientation of the camera. Estimated for the current pose. Can be modified using the Euler angle representation (rotation angles) of the matrix.

**3:** Translation vectors used to determine the relative position of the camera Estimated for the current pose.

**4:** Distortion parameters, they model and correct the distortion caused by imperfections in the optical system of the camera.

# Arena definition

An arena is a closed and controlled area, where a tracking experiment will take place, the arena is separated physically from the outside and from other tracking arenas.

Additionally, inside each arena, the tracking area must be defined prior to the execution. The tracking area should be a white uniform well illuminated area where the objects of interest will be studied. The objects of interest should appear as dark areas with high contrast inside the tracking area.

It is important than areas outside the white uniform region where the objects of interest are located are not inside the tracking areas or present also a white and uniform color. To void phenomena such as external objects, reflections, bubbles, shadows or interferences with walls and corners are ruled out.

The arena definition algorithm, provide a semiautomatic and easy to use tool to define and visualize the arenas and tracking areas.

## 1. Arena definition algorithm

### Automatic selection

The arena selection algorithm takes a sample image of from an input video sequence to define the different tracking areas.

The algorithm to define arenas and tracking areas proceeds as follows:

- The user selects one frame of the video sequence to use as reference of for the algorithm.
- Preprocessing: After the image has been obtained, the calibration model is used to create a distortion map for every pixel of the image, and then interpolation is used to create a distortion-free image. Then image is converted to a 8-bit grayscale, and normalized to value of 0-255.
- Segmentation: The objective of this operation is to separate the tracking areas, which by definition are uniform bright region of the images. First, an intensity value (*roi.thre*) is selected by the user to threshold the image into two binary sets. Finally, a closing mathematical morphological operation is executed. In mathematical morphology, the closing of a binary image *A* by a structuring element *B* is the erosion of the dilation of that set. This operation removes the holes and imperfections on the area selected by the previous operations. The size of the structuring element (*roi.elms*) and the iterations of the dilation (*roi.dilt*) and erosion (*roi.erot*) operations can be selected by the user.
- Arena and area creation: First, the areas obtained in the previous step al filtered according to its size (*roi.mins*). Then, the resulting areas which possess an arbitrary shape are approximated by a polygon so that the number of vertices is the smaller possible and the distance between vertices is less or equal to the precision specified by the user (*roi.poly*). This step simplifies the shape of the selected area, and it help to eliminate some irregularities at the edges. These polygons will constitute the tracking areas. Finally, arenas are defined as the minimum rectangular regions in the image containing each one of the tracking areas.

In the application each arena will be defined as image region, containing a polynomial tracking area. Arenas are obtained by a map describing the undistorted positions of its pixels. Each arena constitutes unit processed in parallel with an independent tracking algorithm.

### Manual selection

The arena selection algorithm takes a sample image of from an input video sequence to define the different tracking areas.

The algorithm to define arenas and tracking areas proceeds as follows:

- The user selects one frame of the video sequence to use as reference of for the algorithm.
- Preprocessing: After the image has been obtained, the calibration model is used to create a distortion map for every pixel of the image, and then interpolation is used to create a distortion-free image. Then image is converted to a 8-bit grayscale, and normalized to value of 0-255.
- Arena creation: The user manually draws an arbitrary number of rectangular shapes, which will constitute the arenas. Each arena constitutes unit processed in parallel with an independent tracking algorithm and each arena will contain a tracking area where the animals will be visible.
- Segmentation: First, an intensity value (*roi.thre*) is selected by the user to threshold the image into two binary sets. Finally, a closing mathematical morphological operation is executed. In mathematical morphology. This operation removes the holes and imperfections on selected area. The size of the structuring element (*roi.elms*) and the iterations of the dilation (*roi.dilt*) and erosion (*roi.erot*) operations can be selected by the user. The largest selected area inside each arena defined by the user is selected.
- Area creation: The largest selected area inside each arena defined by the user is selected. The user has the option to fit each of these areas to a circular shape (*roi.fite*), defined by the minimum enclosing circle containing these areas, and the user can reduce the radius of the circle, by a selected number of pixels (*roi.redr*).

## 2. Arena definition screen

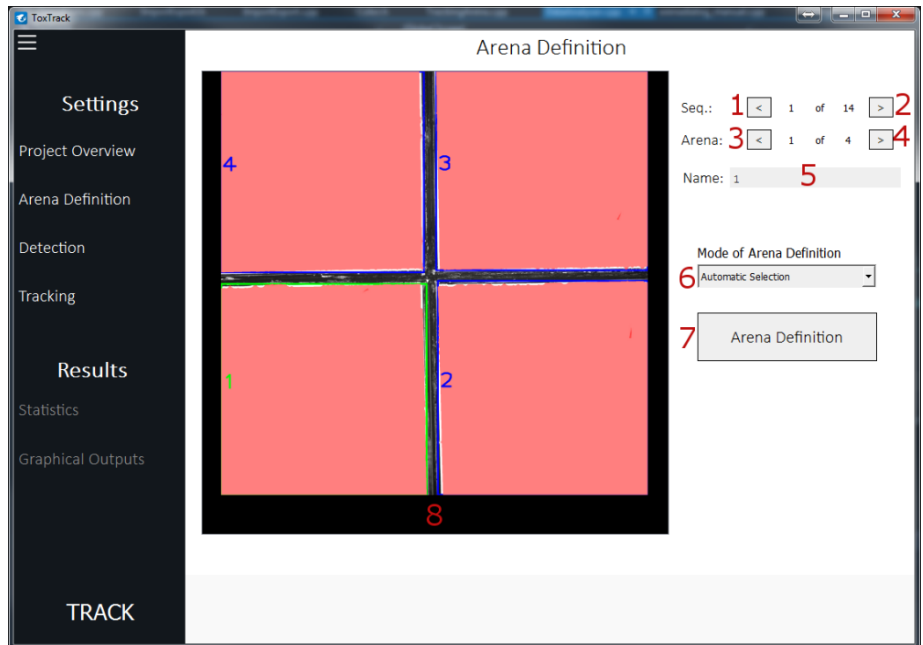

Figure 8. Arena definition screen.

**1:** Select previous sequence. (Allows to preview the results of the current parameters in all the video sequences)
**2:** Select next sequence.
**3:** Select previous arena. The arena currently selected is displayed as a green rectangle, and other arenas are displayed in blue.
**4:** Select next arena.
**5:** Allows the user to introduce a name for the current arena.
**6:** Select arena definition algorithm. (Automatic selection or manual selection).
**7:** Open arena definition window, for the selected algorithm.
**8:** View of the current arena selection. Shows a preview of the current video sequence and highlights the current arena. The tracking areas are displayed in red, and the arenas are shown as colored rectangles with the name over imposed.

## Automatic selection

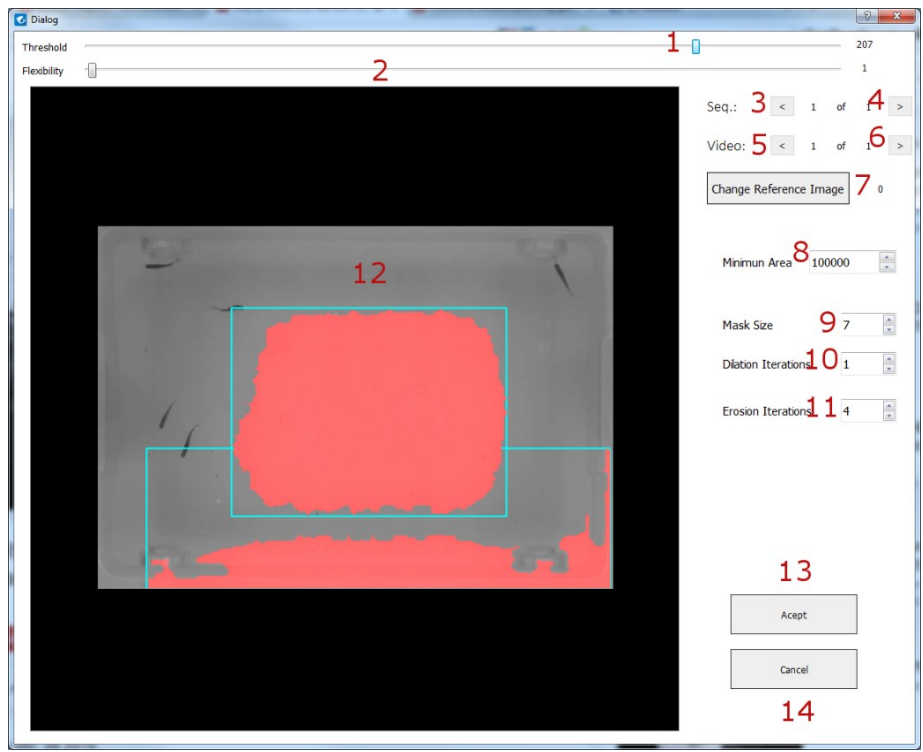

Figure 9. Arena definition screen, automatic selection.

**1:** Threshold, this parameter selects the minimum intensity level (normalized to 1-255) of the tracking areas.

**2:** Polygon Flexibility, defines the maximum distance between vertices in the polygon with the smaller possible number of vertices used to define the tracking areas.

**3:** Select previous sequence. (Allows to preview the results of the current parameters in all the video sequences)

**4:** Select next sequence.

**5:** Select previous video of the current sequence.

**6:** Select next video of the current sequence.

**7:** Change the current frame of the selected video (when the button is pressed a new random frame of the video will be selected and displayed), the selected frame will be used for the selected sequence in the arena definition algorithm.

**8:** Minimum area, minimum number of pixels to constitute a tracking area.

**9:** Mask Size, size (diameter) of the structuring element in the closing mathematical morphology operation.

**10:** Dilation Iterations, number of dilation operation in the closing operation.

**11:** Erosion iterations, number of erosion operation in the closing operation.

**12:** View of the current arena selection. The tracking areas are displayed as red areas, and the arenas as blue rectangles.

**13:** Accept the arena selection.

**14:** Cancel, closes the window, no changes will be made.

## Manual selection

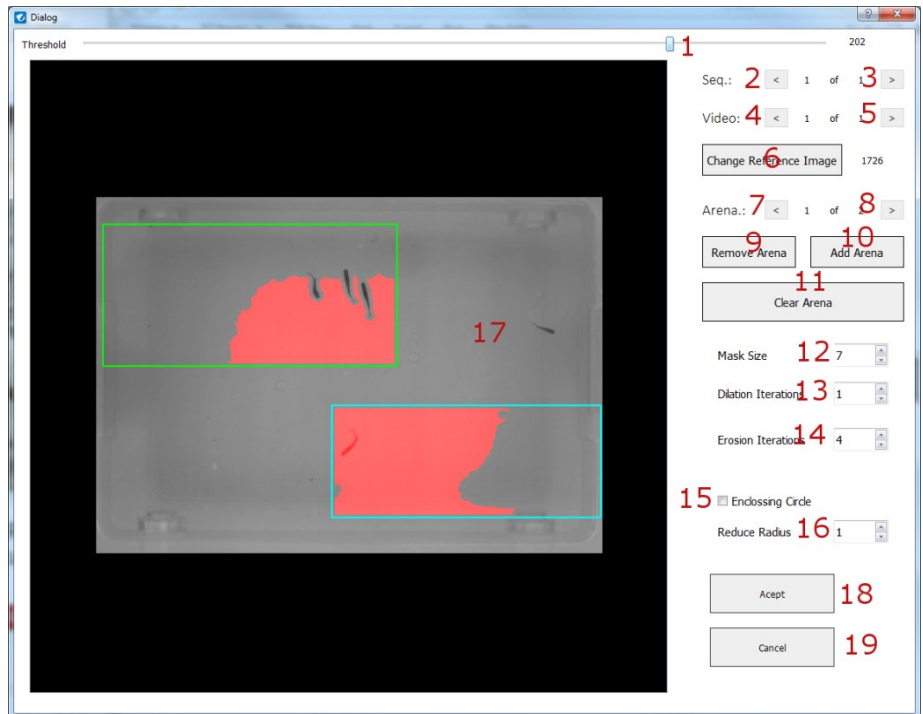

Figure 10. Arena definition screen, manual selection.

**1:** Threshold, this parameter selects the minimum intensity level (normalized to 1-255) of the tracking areas.

**2:** Select previous sequence. (Allows previewing the results of the current parameters in all the video sequences)

**3:** Select next sequence.

**4:** Select previous video of the current sequence.

**5:** Select next video of the current sequence.

**6:** Change the current frame of the selected video (when the button is pressed a new random frame of the video will be selected and displayed), the selected frame will be used for the selected sequence in the arena definition algorithm.

**7:** Select previous arena. The arena currently selected is displayed as a green rectangle, and other arenas are displayed in blue.

**8:** Select next arena.

**9:** Remove selected arena.

**10:** Add arena, opens a drawing window, which can be freely resized, and where the user can draw a rectangular area with the mouse. The drawing window can be closed pressing in the corner of the window or pressing enter.

**11:** Remove all arenas.

**12:** Mask Size, size (diameter) of the structuring element in the closing mathematical morphology operation.

**13:** Dilation Iterations, number of dilation operation in the closing operation.

**14:** Erosion iterations, number of erosion operation in the closing operation.

**15:** Fits the tracking areas to their minimum enclosing circles.

**16:** If the tracking areas are fitted to their minimum enclosing circles. The radius of the circles can be reduced by an arbitrary amount.

**17:** View of the current arena selection. The tracking areas are displayed as red areas, and the arenas as blue or green rectangles.

**18:** Accept the arena selection.

**19:** Cancel, closes the window, no changes will be made.

# Detection

## 1. Detection algorithm

This algorithm will detect dark moving animals in a bright homogenous background. If after the arena selection procedure, the tracking area still contains static objects of a significant size, they can be removed using the background subtraction technique. However, this technique may only be used if the animals do not remain stationary (especially at the beginning of the video).

### Background subtraction

The background subtraction process removes the static elements of the image, using a dynamic background modelling technique. According to this, every pixel of the scene must be matched to the background or foreground category. To this end a widely used model based in estimate the RGB color space probability distribution for every pixel in the image has been chosen. The segmentation algorithm works using Bayesian probability to calculate the likelihood of a pixel $x_{ij}$, at time $t$ in coordinates *(i,j)*, being classified as background *(BG)* or foreground *(FG)*. This is expressed as follows:

$$p(BG \mid x) = \frac{p(x \mid BG) p(BG)}{p(x \mid BG) p(BG) + p(x \mid FG) p(FG)}, \tag{5}$$

In a general case, we can assume that we don't know about the foreground objects and we may assume that *p(BG)=p(FG)=0.5* or we can use a different probability (*bgs.ratb*) according to the knowledge of the scene. The background model will be referred as $p(x \mid BG)$ and we will decide that the pixel belongs to the background if $p(x \mid BG)$ is higher than a certain level.

The background model will be estimated from a set of observations $\chi_{ij} = \left\{ x_{ij}(t-T), ..., x_{ij}(t) \right\}$ where $T$ (*bgs.nums*) is a time period used to adapt to changes. For each new sample, we update the training data set. According to this, new samples are added to the set and old ones are discarded, while the set size does not exceed a certain value.

We will model distribution of a particular pixel as a mixture of Gaussians following the technique proposed in[3,4]. Pixel values that do not fit the background distribution are considered foreground until there is a Gaussian that includes them with sufficient, evidence of supporting it. The $M$ (*bgs.numg*) Gaussian mixture models can be expressed as:

$$p(x \mid \chi, BG + FG) = \sum_{m=1}^{M} \omega_m \mathcal{N} \left( \tau^2{}_m I \right), \tag{6}$$

where $\mu_m$ and $\sigma^2{}_m$ are the estimates of the mean and variance that describe the Gaussian component $m$. The covariance matrices are assumed to be diagonal and the identity matrix $I$ has proper dimensions. Finally, $\omega$ are positive mixing weights that add up to 1.

Given a new data sample $x^t$ at the time *t*, the recursive update equations are:

$$\omega_m \leftarrow (1-\alpha)\omega_m + \alpha o_m^t, \tag{7}$$

$$\mu_m \leftarrow (1-p_m^t)\mu_m + p_m^t x^t, \tag{8}$$

$$\sigma^2_m \leftarrow (1-p_m^t)\sigma^2_m + p_m^t \left( (x^t - \mu_m)^T (x^t - \mu_m) \right), \tag{9}$$

$$p_m^t = \frac{\alpha o_m^t}{\omega_m}, \tag{10}$$

where $\alpha$ (*bgs.lstp*) describes a exponentially decaying envelope which is used to limit the influence of the old data and being approximately $\alpha = 1/T$. The ownership $o_m^t$ is set to one for the "close" component with the largest $\omega$. We define that a sample is "close" to a component if the *Mahalanobis* distance from the component is, less than a particular value (*bgs.thre*). If there are no "close" components a new one is generated and if the maximum number of components is reached, we discard the component with smallest $\omega$.

Usually, the foreground objects will be represented by some additional clusters with small weights $\omega$. Therefore, we can approximate the background model by the first $B$ largest clusters:

If the components are sorted to have descending weights, we have:

$$B = \arg\min_b \left( \sum_{m=1}^{b} \omega_m > (1 - c_f) \right), \tag{11}$$

where $c_f$ is the maximum portion of the data that can belong to the foreground objects without influencing the background model.

### Animal detection

The algorithm to detect the animals in the tracking areas is defined as follows:

- Preprocessing: After the image has been obtained, the distortion map obtained in the calibration is used to create a distortion-free image of the arena. Then the background model removes the static parts of the image. The areas of the image outside the tracking areas defined by the user are also removed from the image.
- Segmentation: An intensity value (*det.thre*) is selected by the user to threshold the image into two binary sets. Finally, a closing mathematical morphological operation is executed. This operation removes small holes and imperfections detected bodies. The size of the structuring element (*det.elms*) and the iterations of the dilation (*det.dilt*) and erosion (*det.erot*).
- Filtering: The objects smaller or bigger than the minimum (*det.mins*) and maximum (*det.maxs*) size limits defined by the user are disregarded.

### Filtering

Additionally to the minimum (*det.mins*) and maximum (*det.maxs*) size limits defined by the user.
The user can also define a number of additional operations and filters listed as follows:

- A closing morphological operation can be used setting the number of dilation operations (*det.dilt*), the number of erosion operations (*det.erot*), and the size of the structuring element used (*det.elss*).

- Objects can be filtered according to the radius of the corresponding minimum enclosing circle, setting a minimum (*det.minr*) and a maximum (*det.maxr*) size limit.
- Objects can be filtered according to the rate between the major and the minor radius of the minimum ellipse fitting the detected body, setting a minimum (*det.mish*) and a maximum (*det.mash*) size limits.
- Objects can be filtered according to the ratio between the area of the minimum ellipse fitting the body and the actual number of pixels detected in the body, setting a minimum threshold indicating the minimum fill rate of the object (*det.minf*).

## 2. Detection screen

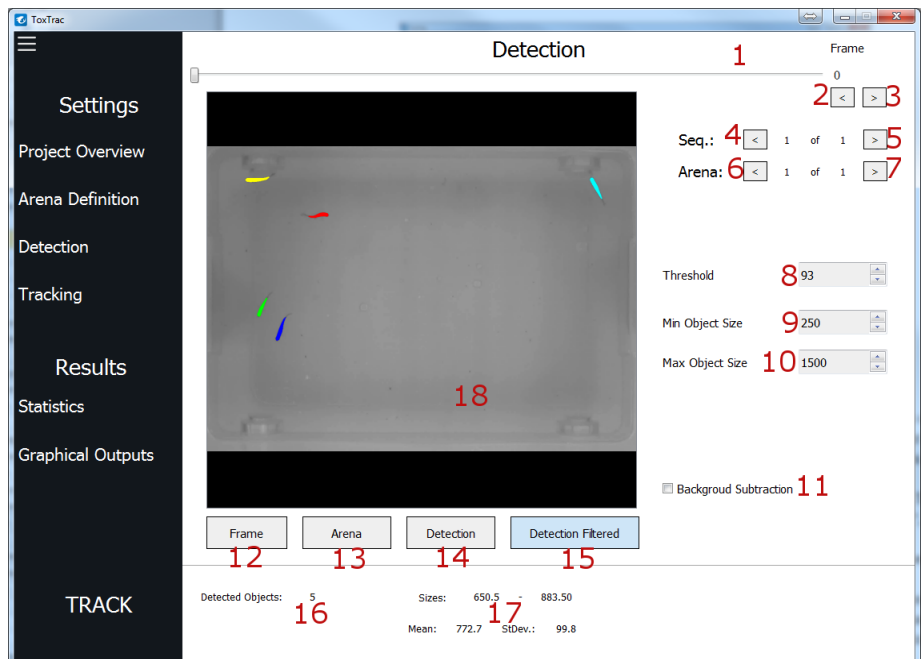

Figure 11. Detection screen.

**1:** Navigate through the frames of the video sequence.
**2:** Displays previous frame.
**3:** Displays next frame.
**4:** Select previous sequence.
**5:** Select next sequence.
**6:** Select previous arena.
**7:** Select next arena.
**8:** Threshold, this parameter selects the minimum intensity level (1-255) for the detected objects.
**9:** Selects minimum object size in pixels.
**10:** Selects maximum object size in pixels.
**11:** Allows to enable or disable the background subtraction algorithm (subtracted background will not be shown in the visualization window).
**12:** Shows the original video frame.
**13:** Shows the areas of the image excluded from the tracking area.
**14:** Shows unfiltered detections.
**15:** Shows filtered detections.
**16:** Shows the number of detected objects with the current parameters.

**17:** Shows the range of sizes of de detected objects, together with the mean and standard deviation of the sizes.

**18:** Displays current detection in the selected arena and video frame. Calibration distortion is applied to the displayed image. Detected objects are displayed in different colors over imposed to the image. The user should check that all animals are detected with the current parameters, and no objects of the background are detected.

# Tracking

## 1. Kalman tracking algorithm

Tracking is the problem of generating an inference about the motion of one or more objects from a sequence of images. The Kalman filter addresses the problem of estimating the state $x \in R^n$ of a discrete-time controlled process that is governed by the linear equation, expressed as follows:

$$x(t+1) = Ax(t) + w(t),\tag{12}$$

where $A$ is a $n$ by $n$ matrix called state transition matrix, which relates the state of the system at the previous time step to the state at the current step, and $w$ represents the process noise, which is assumed normally distributed with mean 0.

For the state transition matrix, we consider the equations of two-dimensional motion assuming a constant acceleration between time steps:

$$x_{t+1} = x_t + v_{x,t} + \frac{1}{2}a_x,\tag{13}$$

$$y_{t+1} = y_t + v_{y,t} + \frac{1}{2}a_y,\tag{14}$$

$$v_{x,t+1} = v_{x,t} + a_x,\tag{15}$$

$$v_{y,t+1} = v_{y,t} + a_y,\tag{16}$$

where $(x, y)$ is the animal position, $(v_x, v_y)$ is the velocity and $(a_x, a_y)$ is the acceleration, which is assumed constant in a time step. We also consider an observation model described by the following equation:

$$z(t) = Hx(t) + v(t),\tag{17}$$

where $z \in R^m$ represents the measurement, $H$ is a $m$ by $n$ matrix called observation matrix and $v$ is the measurement error, which is assumed independent of $w$ and normally distributed with mean 0.

So the model equations can be expressed as follows:

$$\begin{bmatrix} x_{t+1} \\ y_{t+1} \\ v_{x,t+1} \\ v_{y,t+1} \end{bmatrix} = \begin{bmatrix} 1 & 0 & t & 0 \\ 0 & 1 & 0 & t \\ 0 & 0 & 1 & 0 \\ 0 & 0 & 0 & 1 \end{bmatrix} \begin{bmatrix} x_t \\ y_t \\ v_{x,t} \\ v_{y,t} \end{bmatrix} + w(t),\tag{18}$$

$$\begin{bmatrix} z_x \\ z_y \end{bmatrix} = \begin{bmatrix} 1 & 0 & 0 & 0 \\ 0 & 1 & 0 & 0 \end{bmatrix} \begin{bmatrix} x_t \\ y_t \\ v_{x,t} \\ v_{y,t} \end{bmatrix} + v(t) \,, \tag{19}$$

The Kalman filter works in a two-step recursive process. First, it estimates the new state, along with their uncertainties. Once the outcome of the next measurement (corrupted with noise) is observed, these estimates are updated. The algorithm can, run in real time using only the current input measurements and the previously calculated state. In the present work, the implementation of the Kalman filter was performed according to[5], using an empirical estimate of the measurement error and the process noise covariances.

The Kalman filter essential problem is the assignment of detections to tracks. To this end, a cost is assigned to every possible pair of track–detection. The cost is understood as the probability of that detection to correspond to the current track position. It is calculated using the distance from the detected position to the predicted position of the animals. To this end, the minimum of the Euclidean distances is selected as cost metric according to the Hungarian optimization algorithm[6].

To create a Kalman filter, we need to initialize the noise covariance matrices $v$ and $w$, and the a posteriori error covariance matrix $p$. To this end we have used the following procedure.

$$w = \begin{bmatrix} dt^4/4 & 0 & dt^3/2 & 0 \\ 0 & dt^4/4 & 0 & dt^3/2 \\ dt^3/2 & 0 & dt^2 & 0 \\ 0 & dt^3/2 & 0 & dt^2 \end{bmatrix} a \,, \tag{20}$$

$$v = \begin{bmatrix} m & 0 \\ 0 & m \end{bmatrix}, \tag{21}$$

$$p = \begin{bmatrix} c & 0 & 0 & 0 \\ 0 & c & 0 & 0 \\ 0 & 0 & c & 0 \\ 0 & 0 & 0 & c \end{bmatrix}, \tag{22}$$

where $dt$ ($kal.time$) represent a time increment magnitude, $a$ ($kal.pron$) represent the estimated process noise, $m$ ($kal.mean$) represent the estimated measurement noise, and $c$ ($kal.errc$) is a value used to initialize the a posteriori error covariance matrix with a correct value. Modifying the $a$ and $m$ values will change the behavior of the filter, causing to use more aggressive predictions (not recommended) or to stick more to the measured positions. These values can be modified by user in the configuration text file.

## Feature Extraction

To be able to find the identity of individuals when multiple objects are tracked and there is risk of occlusion, we calculate a set of characteristic features for a detected body $B$. These features are the intensity histogram (HIST) and two Texture Center Maps (TCM) called the Intensity Center Map (ICM); and the Contrast Center Map (CCM), where the latter two are 2D distributions. They are defined as:

$$HIST_B(x) = \left| \left\{ p : f_p = x \right\} \right|, \tag{23}$$

$$ICM_B(x,y) = \left| \left\{ p : \|p-c\| = x, f_p + f_c = y \right\} \right|, \tag{24}$$

$$CCM_B(x,y) = \left| \left\{ p : \|p-c\| = x, f_p + f_c = y \right\} \right|, \tag{25}$$

where $p, c \in B$, being $p$ an arbitrary pixel and $c$ the center of mass of the body $B$. $f_p$ and $f_c$ represent the color values of $p$ and $c$ respectively, and $\|p-c\|$ is the Euclidean distance between $p$ and $c$. Therefore, for a detected body in an image, we define the detection $d$ as:

$$d(B) = \left\{ c, \left| \{p\} \right|, HIST_B, ICM_B, CCM_B, T : p, c \in B \right\}, \tag{26}$$

where $T$ is the time, $\left| \{p\} \right|$ the size, and $c$ is the position of the body.

The way this features are created and stored by the program can be altered. Therefore, the number of clusters of the histogram *(kal.hiss)*, the maximum distance from the body center used by the TCM maps *(kal.tcmr)*, the data type used to store the TCM maps in memory *(kal.tcmd)*, and the maximum number of features stored for a track *(kal.hist)*.

## Acceptance conditions

In order to reduce the possibility of assigning the wrong detection to a track, we have implemented an additional step to check if there is a significant change in the position or in the in object size of the detections.

Assuming that the detection $d_i$ has been assigned by the Hungarian algorithm to the track $t_j$, formed by the detections $t_j = \{d_{J0}, ..., d_{Jn}\}$, the following parameters are calculated:

**Frame distance (c1)**: The distance from a detection $d_j$ to a track $f_i$, is defined as the Euclidean distance from the predicted position of the track $d_{Jn+1}$ to $d_i$.

$$c1 = \|t_j - d_i\| = \|d_{Jn+1} - d_i\|, \tag{27}$$

**Size change (c2):** The size of the detected body in $d_i$ is compared with all detections in $f_j$, and the minimum relative change is selected

$$c2 = \max\left( \left( |d_i| - |d_{Jn}| \right) / |d_{Jn}|, \left( |d_{Jn}| - |d_i| \right) / |d_i| \right), \tag{28}$$

To accept the assignment of $d_i$ to $t_j$ the following conditions must occur:

$$\begin{aligned} &Accept \leftarrow c2 < S_H \ and \ c1 < d \times (T_i - T_{Jn}) \\ &Accept \leftarrow c2 < S_H \ and \ c1 < d \end{aligned}, \tag{29}$$

where $d$ *(kal.disf)* represent a frame distance condition, $S_H$ *(kal.sich)* represent a size change condition.

## Collision

As explained in the main document, when two individuals overlap or cross. We mark the conflicted tracks as inactive tracks and generate new ones. To detect a collision we define the operator $cl(t_j) = d_i$ to refer to the closest detection $d_i$ to the track $t_j$.

A track is classified as conflicted if one of the following conditions applies:

$$cl(t_i) = cl(t_k) = d_i \text{ and } abs(\|t_k - d_i\| - \|t_k - d_i\|) < (d \times a_d), \tag{30}$$

$$abs(\|t_k - d_i\| - \|t_k - d_i\|) < a, \tag{31}$$

where $a_d$ (*kal.advr*) is defined from 0 to 1, and $a$ (*kal.advm*) represent a minimum advantage value in pixels.

## Delete conditions

To avoid that occasional missdetections interfere with the final results, tracks that accomplish one of the following conditions are deleted.

- Tracks which are non-active and are smaller than the minimum age (*kal.dmax*). If the size of the track is smaller than (*kal.mins*) the track is also marked as short.
- Tracks not assigned to any detection for a certain amount of frames (*kal.dage*) are marked as inactive. If the size of the track is smaller than (*kal.mins*) the track is also marked as short.

## 2. Fragment Identification

This is an optional post-processing step that calculates what trajectory fragments that belong to each individual, preserving the identity of the animals after an occlusion. To assign the correct identities, we compute and study the similarities of the trajectory features. First, an identity matrix is constructed, containing a similarity value for each pair of tracks $Sim(t_{row}, t_j)$. This value is constructed comparing the stored features of those tracks, and represent the likelihood of those tracks to correspond to the same animal.

Since to estimate the similarity of two tracks is a very computationally expensive procedure, only the samples similar in size *(kal.cmsc)*, and with a minimum histogram correlation value (*kal.cmhc*) are used to construct the similarity values $Sim(t_{row}, t_j)$

To assign the different tracks to each other, the first step is to select groups of long tracks, coexisting at the same time and belonging to all individuals. We know this is an assignment problem with an optimal solution, so we use a variant of the classic Hungarian optimization algorithm (Kuhn 1955). The minimum similarity value (*kal.idgb*) accepted by the algorithm can be modified by the user.

To assign the remaining tracks, we iteratively select the best correlation value in the *matrix*, first with the long and then with the short tracks. The minimum similarity values (*kal.idlb, kal.idsb*) and the minimum average similarity values (*kal.idla, kal.idsa*) accepted can also be modified by the user.

With every assignment, we update the matrix propagating the knowledge obtained iteration and reducing the uncertainty for the remaining tracks.

Since the fragment identification algorithm works as a post processing technique, and requires a big amount of memory to store the features of active and inactive tracks, the number of tracks kept in memory can be defined by the user (*kal.idff*). When the number of tracks reaches this number, the algorithm will try to identify the tracks to release memory, before continuing the tracking

## 3. Tracking screen

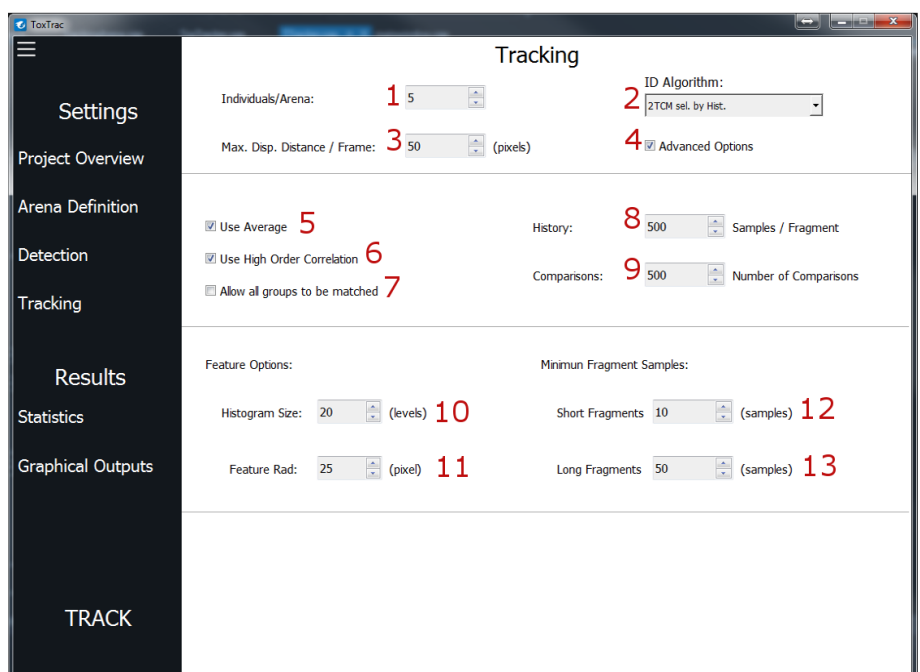

Figure 12. Tracking Screen.

**1:** Select the number of animals in each arena. (All arenas should have the same number of animals, or be empty).

**2:** Selects fragment identification algorithm. Available options are: **"*Not Id*"** (no fragment identification algorithm will be used) **"*Hist. sel. by Shape (beta)*"** (only histogram information will be used in the fragment identification algorithm. Reduces processing time and memory use significantly but this algorithm is still under development), and **"*2TCM sel. by Hist.*"** (uses the fragment identification algorithm proposed in the paper).

**3:** Selects the maximum pixels per frame, an animal is allowed to move. It is recommended to overestimate this parameter.

**4:** Enables the user to change advanced options (5-16).

**5:** Use Avg. Use a mean metric instead of a max similarity value (Recommended).

**6:** Use High order Correlation. Takes in account the similarity with all fragments, without reducing the speed. (Recommended).

7: Allow all groups to be matched. Increases accuracy, but may lead to have a final number of tracks different to the number of individuals.

**8:** History, maximum number of samples used for every track in the algorithm. This parameter also affects significantly the memory and time used by the fragment identification algorithm.

**9:** Comparisons, limits the number of comparisons between two samples of the same track in order to increase speed.

**10:** Size (number of color clusters) used by the histograms.

**11:** Max distance from the body center used by the TCM maps. Setting a higher the value, will allow to use a greater part of the animal bodies in the algorithm, increasing accuracy, memory use and required time.

**12:** Minimum track size to be used by the algorithm, smaller tracks will not be saved.

**13:** Minimum track size to classify it as long.

# Results

## 1. Statistics screen

### Individual statistics

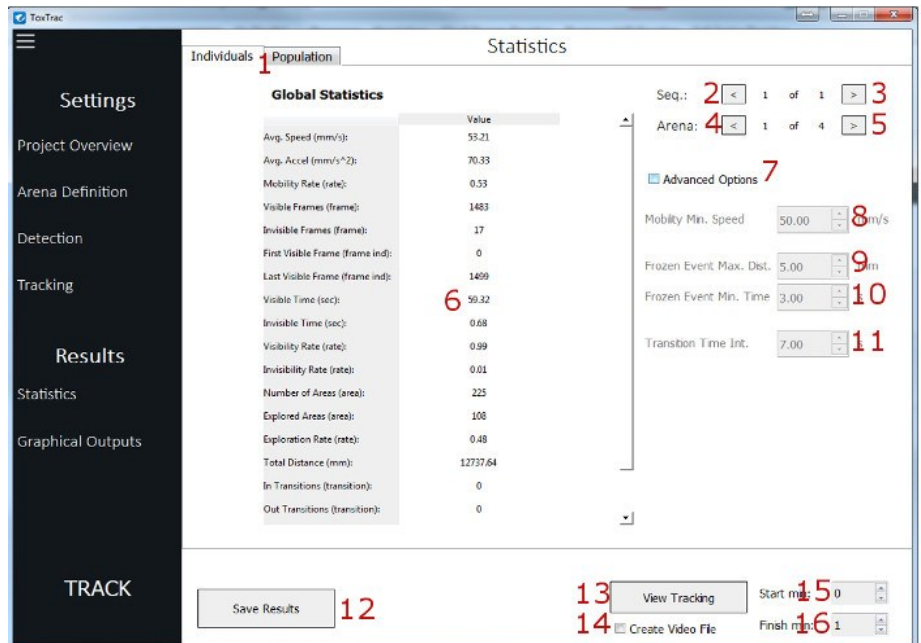

Figure 13. Statistics screen, individuals.

**1:** Change between information of one single individual and for the entire population.

**2:** Select previous sequence.

**3:** Select next sequence.

**4:** Select previous arena.

**5:** Select next arena.

**6:** Displays the tracking statistics for the selected arena and sequence.

**7:** Enables the user to change advanced options (8-11).

**8:** Allows the user to change the speed threshold to estimate the mobility rate of the animal in the stats.

**9:** Allow the user to change the distance threshold (in mm) to calculate the frozen events. A frozen event is detected when the animal has moved less than this value in time interval.

**10:** Allow the user to change the time threshold (in seconds) to calculate the frozen events. A frozen event is detected when the animal has moved less than a distance threshold during a lapse of time.

**11:** Allow the user to change the time threshold (in seconds) to calculate a transition. A transition is detected when the time between two consecutive detections exceed this value, signaling a period of time when the animal is not visible.

**12:** Saves all results in the output folder.

**13:** Shows a free resizable window showing the tracking for the current arena, the window can be closed pressing the 'esc' key, of clicking in the window upper right corner.

**14:** If enabled the generate output button will also save a video of the tracking in the output folder.

**15:** Select the starting point (for all sequences) to show the tracking.

**16:** Select the ending point (for all sequences) to show the tracking.

## Population statistics

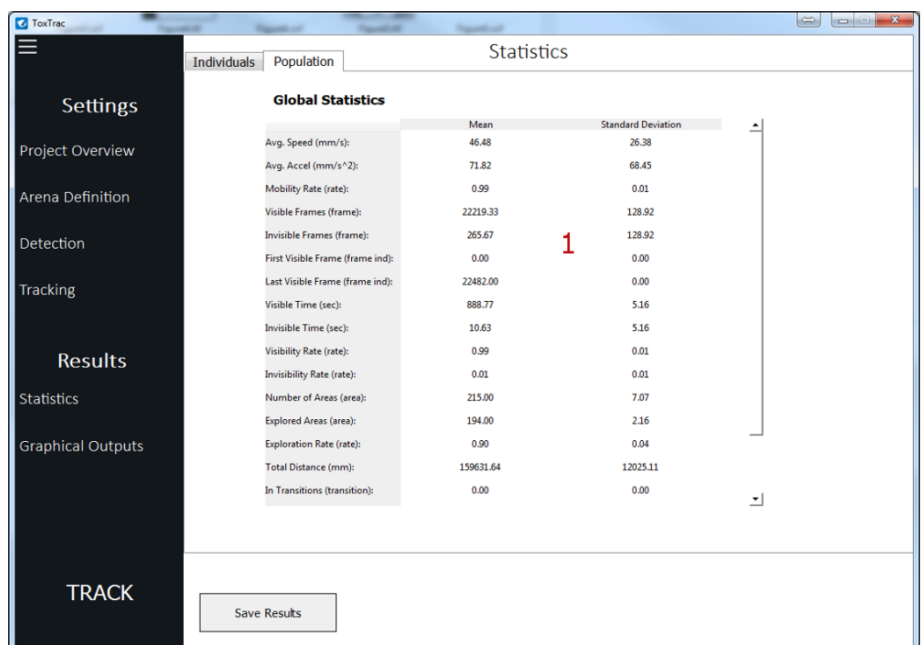

Figure 14. Statistics screen, population.

**1:** Displays the tracking statistics for the entire population.

## 2.  Graphical outputs screen

### Individual graphical outputs

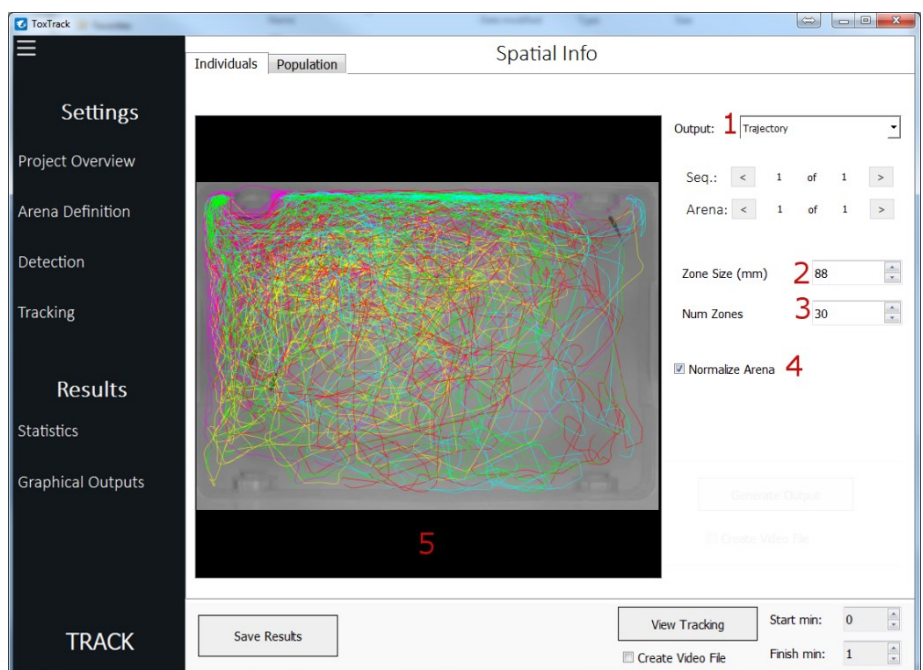

Figure 15. Graphical outputs, individuals.

**1:** Selects one of the different graphic available results. The graphical outputs will be shown superimposed to the arena picture, and in real scale. The heat maps are color coded according to the normalized representation of the frequency of use, in the linear scale shown in Figure 19.

Available options are:

- **_"Plain Output"_**. This option will show the first frame of the sequence and arena without any graphical output.
- **_"North Edge"_**. This option will show a heat map showing the use of different zones in the arena according to the distance of to the North (N) wall. See section: "Distance to Edges (*Dist_Edges.txt, Dist_Edge_[edge].jpeg*)".
- **_"West Edge"_**. This option will show a heat map showing the use of different zones in the arena according to the distance of to the North (W) wall. See section: "Distance to Edges (*Dist_Edges.txt, Dist_Edge_[edge].jpeg*)".
- **_"South Edge"_**. This option will show a heat map showing the use of different zones in the arena according to the distance to the North (S) wall. See section: "Distance to Edges (*Dist_Edges.txt, Dist_Edge_[edge].jpeg*)".
- **_"East Edge"_**. This option will show a heat map showing the use of different zones in the arena according to the distance to the North (E) wall. See section: "Distance to Edges (*Dist_Edges.txt, Dist_Edge_[edge].jpeg*)".
- **_"All Edges"_**. This option will show a heat map showing the use of different zones in the arena according to the distance to any wall. See section: "Distance to Edges (*Dist_Edges.txt, Dist_Edge_[edge].jpeg*)".
- **_"Exploration"_**. This option shows a heat map showing the use of different zones in the arena using a regular grid with a grid of a size selected by the user See section; "Exploration (*Exploration.txt, Exploration.jpeg*)".
- **_"Center Point Dist."_**. This option will show a heat map showing the use of different zones in the arena according to the distance to the center of the arena. See section: Distance to Center Position (*Dist_CenterPos.txt, Dist_Center_Pos.jpeg*).
- **_"Mean Point Dist."_**. This option will show a heat map showing the use of different zones in the arena according to the distance to the mean point of the detected animals in the arena. See section: "Distance to Mean Position (*Dist_MeanPos.txt, Dist_Mean_Pos.jpeg*)".
- **_"Trajectory"_**. This option shows a representation of the trajectory of the animals. See section: "Tracking in Real Space coordinates (*Tracking_RealSpace.txt, Trajectory.jpeg*)".

**2:** Change zone zize in mm. This will affect the zone statistics.

**3:** Number of zones. This will affect the edge zone statistics.

**4:** Normalize arenas. If enabled, the zones will be computed according to the extremes of the detected positions instead of using the full image.

**5:** Visualization panel, displays the current graphic selected in the first image of the current video, for the current arena.

## Population graphical outputs

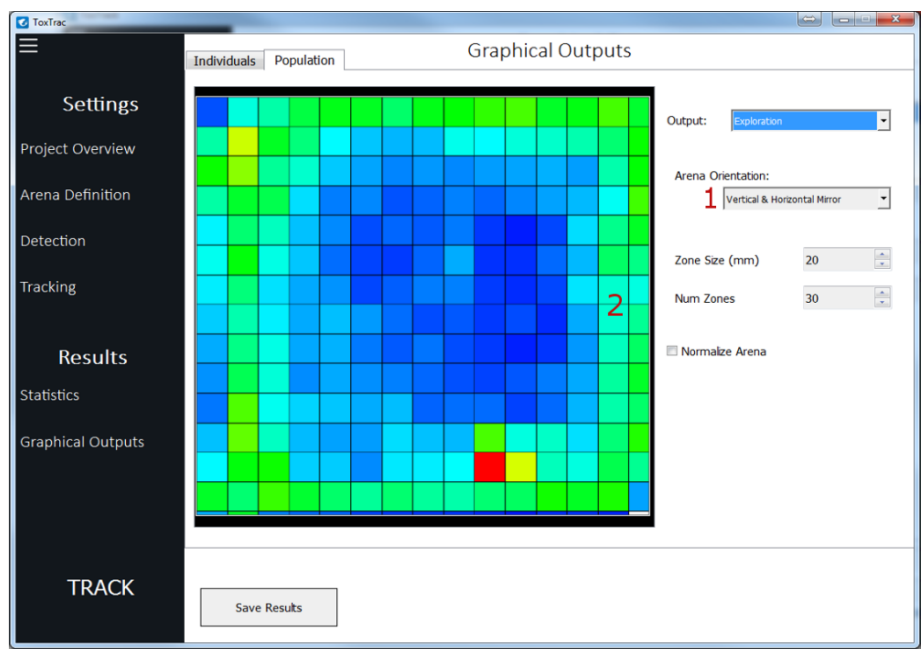

Figure 16. Graphical outputs, population.

**1:** If the arenas are not symmetrical, and do not have the same orientation. This button allows to change the orientation of the arenas, in the projected virtual arena for the entire population. Available options are: **"Same Orientation"**, "**Vertical Mirror"** (the orientation of the arenas in the upper and lower half of the image is inverted in its vertical axis), **"Horizontal Mirror"** (the orientation of the arenas in the left and right half of the image is inverted in its horizontal axis), **"Vertical & Horizontal Mirror"** (combines the previous two options).

**2:** Visualization panel, displays the selected graphic in a real scale virtual arena, projecting together the results of the entire population.

## 3. Output Files

The application automatically generates a folder structure to save the results of the tracking. A set of files are generated to save the project configuration data (Table 3). The main results of the project, cantaining the majority of the statistical data generated for each individual and the population will be formatted in a datasheet (Figure 4). Additionally, a set of files is created with the results for the entire population (Table 5) and another set of files is created with the results for each arena and sequence (Table 6). The creation of most of these files can be enabled or disabled in the *Configuration.txt* file. When analyzing a project, a folder is generated for every video sequence and the corresponding file names are named ending in a number representing the arena number.

Table 3. Structure of the output project files, these files are detailed in Table 2.

| Output Folder | Output Files |
|---|---|
| C:/ProjectFolder/ | pname.tox |
| C:/ProjectFolder/pname/ | pname_Input.txt<br>pname_Configuration.txt<br>pname_Arena.txt<br>pname_ArenaNames.txt<br>pname_Calibrator.txt<br>pname_Output.txt |

Table 4. Datasheet continuing main results of the experiment.

| File Name | Video | Seq | Output Folder | Output Excel File |
|---|---|---|---|---|
| Seq1_0.avi | 1 | 1 | C:/ProjectFolder/ | pname.xls |
| Seq1_1.avi | 2 | | | |
| Seq2_0.avi | 1 | 2 | | |

Table 5. Structure of the output files containing the results for the entire population.

| File Name | Video | Seq | Output Folder | Output TEXT Files | Output JPG Files |
|---|---|---|---|---|---|
| Seq1_0.avi | 1 | 1 | C:/ProjectFolder/pname | Tracking.txt<br>Tracking_RealSpace.txt<br>Instant_Accel.txt<br>Instant_Speed.txt<br>Dist_MeanPos.txt<br>Dist_CenterPos.txt<br>Dist_Edges.txt<br>Exploration.txt<br>Transitions.txt<br>FrozenEvents.txt<br>Stats.txt | Dist_CenterPos.jpeg<br>Dist_Edge_All.jpeg<br>Dist_Edge_E.jpeg<br>Dist_Edge_N.jpeg<br>Dist_Edge_S.jpeg<br>Dist_Edge_W.jpeg<br>Dist_MeanPos.jpeg<br>Exploration.jpeg<br>Trajectory.jpeg |
| Seq1_1.avi | 2 | | | | |
| Seq2_0.avi | 1 | 2 | | | |

Table 6. Structure of the output files containing the results for each arena.

| File Name | Video | Seq | Seq Output Folder | Arena | Output TEXT Files | Output JPEG Files |
|---|---|---|---|---|---|---|
| Seq1_0.avi | 1 | 1 | C:/ProjectFolder/pname /Seq1/ | 1 | Tracking_0.txt<br>Tracking_RealSpace_1.txt<br>Instant_Accel_1.txt<br>Instant_Speed_1.txt<br>Dist_MeanPos_1.txt<br>Dist_CenterPos_1.txt<br>Dist_Edges_1.txt<br>Exploration_1.txt<br>Transitions_1.txt<br>FrozenEvents_1.txt | Dist_CenterPos.jpeg<br>Dist_Edge_All.jpeg<br>Dist_Edge_E.jpeg<br>Dist_Edge_N.jpeg<br>Dist_Edge_S.jpeg<br>Dist_Edge_W.jpeg<br>Dist_MeanPos.jpeg<br>Exploration.jpeg<br>Trajectory.jpeg |
| Seq1_1.avi | 2 | | | 2 | Tracking_1.txt<br>Tracking_RealSpace_2.txt<br>Instant_Accel_2.txt<br>Instant_Speed_2.txt<br>Dist_MeanPos_2.txt<br>Dist_CenterPos_2.txt<br>Dist_Edges_2.txt<br>Exploration_2.txt<br>Transitions_2.txt<br>FrozenEvents_2.txt | |

## Tracking (*Traking.txt*)

This file is used by the program. And contains the animal detected positions. Coordinates are in pixel and relative to the individual after removing distortion. So they cannot be back-projected directly to the original images. Time coordinates are in frames. An example and the meaning of the columns is the following. The arenas are numbered from 0 to n-1, in accordance to the c++ vector indexing.

Table 7. *Traking.txt* files.

| Frame Number | Arena Number | Track number | X-Position | Y-Position |
|---|---|---|---|---|
| 360 | 0 | 1 | 224.634 | 32.2103 |
| 361 | 0 | 1 | 225.131 | 32.2103 |
| 362 | 0 | 1 | 226.737 | 32.2103 |
| 363 | 0 | 1 | 227.434 | 32.2103 |
| 364 | 0 | 1 | 226.624 | 32.2103 |

| Label | Meaning |
|---|---|
| 0 | Predicted Position |
| 1 | Confirmed position |
| 2 | Occluded position |
| 3 | Mirror position |

## Tracking in Real Space coordinates (*Tracking_RealSpace.txt, Trajectory.jpeg*)

This file is the direct translation of the previous file to a real space. Coordinates are expressed in millimeters (typically, though it depends on the calibration unit used) and time in seconds.

In the files named *Tracking_RealSpace_[arena_number].txt* space coordinates will not be referred to the arena, but to a reference point depending on the pose, so different arenas will have different spatial coordinates, corresponding to their relative position in the real space.

In the file related to the population, *Tracking_RealSpace.txt* the space coordinates will be projected to a common virtual arena, according to a selected arena distribution. If the normalized option is selected (*ana.norm*) so the min and x and y coordinates from each arena will correspond to the same x and y coordinates from the virtual arena. The arenas are numbered from 0 to n-1Here, and in consequent results, the arenas are numbered from 1 to n, in accordance with the values shown in the interface.

Table 8. *Tracking_RealSpace.txt* files.

| Time (sec) | Arena | Track | Pos. X (mm) | Pos. Y (mm) | Label |
|---|---|---|---|---|---|
| 15.0414 | 1 | 1 | 65.2423 | 31.7268 | 1 |
| 15.0414 | 1 | 1 | 64.8647 | 31.7615 | 1 |
| 15.0831 | 1 | 1 | 64.1963 | 31.7955 | 1 |
| 15.1248 | 1 | 1 | 63.6246 | 31.7977 | 1 |
| 15.1664 | 1 | 1 | 63.155 | 31.7606 | 1 |
| 15.2081 | 1 | 1 | 62.7215 | 31.6936 | 1 |
| 15.2498 | 1 | 1 | 62.3192 | 31.6089 | 1 |
| 15.2914 | 1 | 1 | 61.9783 | 31.5102 | 1 |
| 15.3331 | 1 | 1 | 61.7018 | 31.4046 | 1 |
| 15.3748 | 1 | 1 | 61.4741 | 31.2961 | 1 |

The graphic outputs, *Trajectory.jpeg* correspond to colored representation of the trajectory of the animal, superimposed to the arena picture, and in real scale. The trajectory of different tracks in the same arena is colored with a different color.

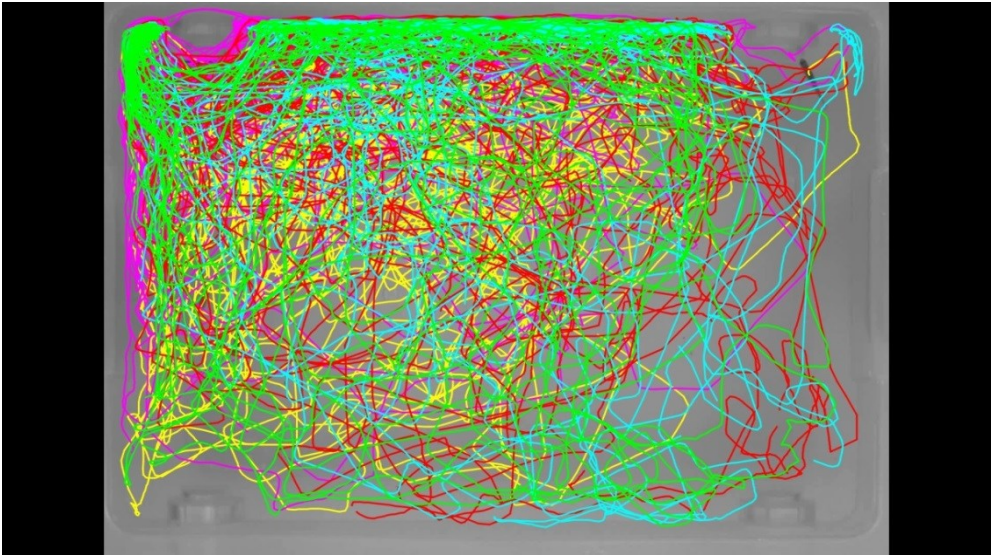

Figure 17. *Trajectory.jpeg.* Trajectory projection of 5 fish in a single arena experimental setup.

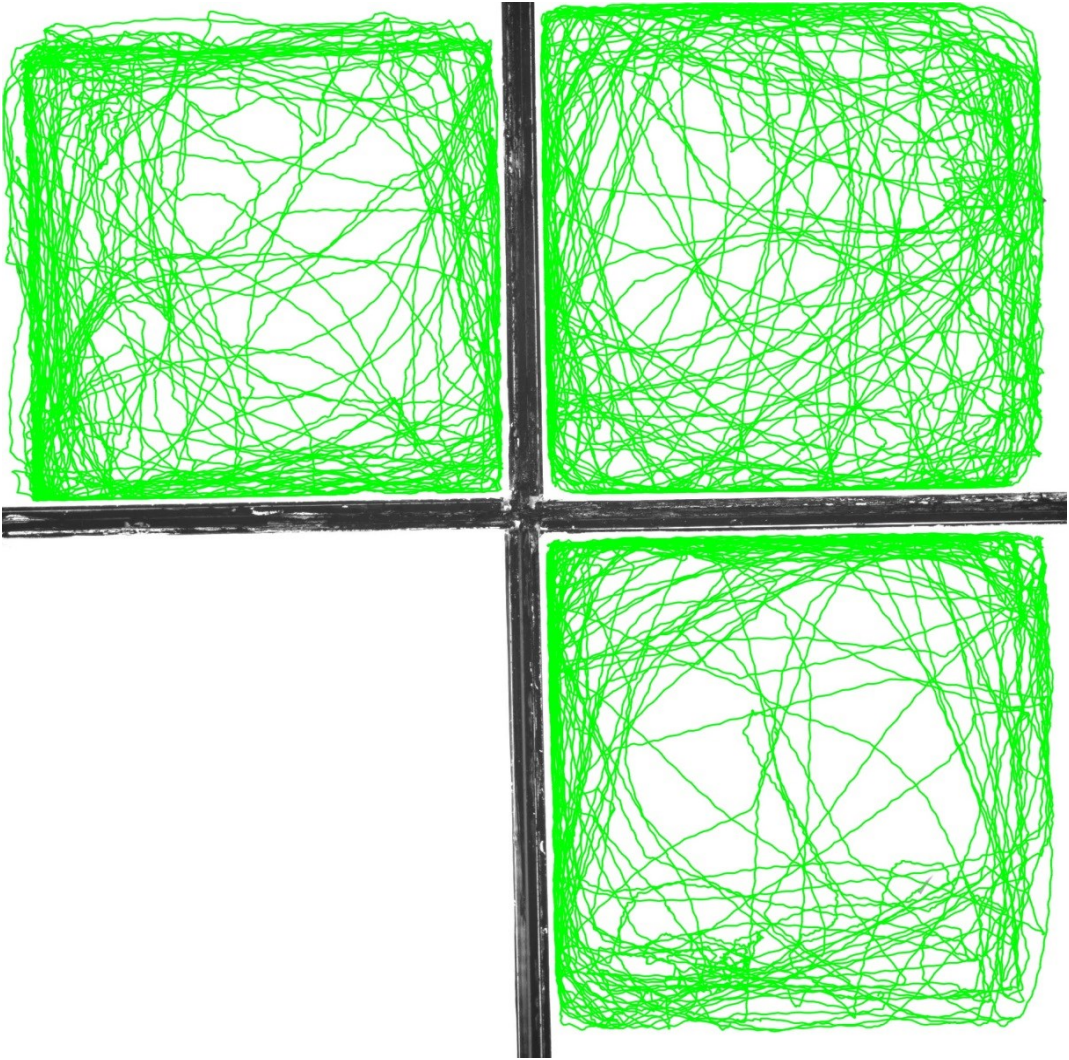

Figure 18. *Trajectory.jpeg.* Trajectory projection of 3 fish in a four arena experimental setup.

## Instantaneous Speed (*Instant_Speed.txt*)

This file is the first derivate of the *Tracking_RealSpace.txt* files. Where the instantaneous speed $s_i$ corresponding to the time $t_i$ is calculated according to the formula:

$$s_i^{arena,track} = \frac{\sqrt{\left(x_{i+c} - x_{i-c}\right)^2 + \left(y_{i+c} - y_{i-c}\right)^2}}{t_{i+c} - t_{i-c}} \,, \tag{32}$$

where $c$ (*ana.spsa*) is a sampling distance defined by the user which can be tuned by the user.

Table 9. *Instant_Speed.txt* files.

| Time (sec) | Arena | Track | Current Speed (mm/sec) |
|---|---|---|---|
| 0.08 | 1 | 1 | 73.2125 |
| 0.12 | 1 | 1 | 59.5829 |
| 0.16 | 1 | 1 | 52.8199 |
| 0.2 | 1 | 1 | 43.0816 |
| 0.24 | 1 | 1 | 32.5418 |
| 0.28 | 1 | 1 | 23.6301 |
| 0.32 | 1 | 1 | 17.503 |
| 0.36 | 1 | 1 | 13.6693 |
| 0.4 | 1 | 1 | 10.7872 |
| 0.44 | 1 | 1 | 13.3477 |

## Instantaneous Acceleration (*Instant_Accel.txt*)

This file is the second derivate of the *Tracking_RealSpace.txt* files. Where the instantaneous acceleration $a_i$ corresponding to the time $t_i$ is calculated according to the formula (accelerations and decelerations will be represented as positive values):

$$a_i^{arena,track} = \frac{\sqrt{\left(s_{i+c} - s_{i-c}\right)^2}}{t_{i+c} - t_{i-c}} \,, \tag{33}$$

where $c$ (*ana.spsa*) is a sampling distance defined by the user which can be tuned by the user.

Table 10. *Instant_Accel.txt* files.

| Time (sec) | Arena | Track | Current Accel. (mm/s^2) |
|---|---|---|---|
| 0.16 | 1 | 1 | 254.192 |
| 0.2 | 1 | 1 | 224.705 |
| 0.24 | 1 | 1 | 220.731 |
| 0.28 | 1 | 1 | 183.827 |
| 0.32 | 1 | 1 | 135.966 |
| 0.36 | 1 | 1 | 64.265 |
| 0.4 | 1 | 1 | 39.7648 |
| 0.16 | 1 | 1 | 254.192 |
| 0.2 | 1 | 1 | 224.705 |
| 0.24 | 1 | 1 | 220.731 |

### Distance to Edges (*Dist_Edges.txt, Dist_Edge_[edge].jpeg*)

We define the walls of the arena by the lines connecting its corner points. To this end we will use a cardinal nomenclature. Therefore the North (N) wall will be defined as the line connecting the North West (NW) and North East (NE) corners of the image and it will correspond to the up side of the arena image.

Therefore, the distance for a detected position $(x_i, y_i)$ corresponding to the time $t_i$, the distance corresponding to the different walls of the arena will be defined as follows:

$$dist_{N,i} = \frac{\left|\left(y_{NE} - y_{NW}\right)x_i - \left(x_{NE} - x_{NW}\right)y_i + x_{NE}y_{NW} - y_{NE}x_{NW}\right|}{\sqrt{\left(x_{NE} - x_{NW}\right)^2 + \left(y_{NE} - y_{NW}\right)^2}}, \tag{34}$$

$$dist_{W,i} = \frac{\left|\left(y_{SW} - y_{NW}\right)x_i - \left(x_{SW} - x_{NW}\right)y_i + x_{SW}y_{NW} - y_{SW}x_{NW}\right|}{\sqrt{\left(x_{SW} - x_{NW}\right)^2 + \left(y_{SW} - y_{NW}\right)^2}}, \tag{35}$$

$$dist_{S,i} = \frac{\left|\left(y_{SE} - y_{SW}\right)x_i - \left(x_{SE} - x_{SW}\right)y_i + x_{SE}y_{SW} - y_{SE}x_{SW}\right|}{\sqrt{\left(x_{SE} - x_{SW}\right)^2 + \left(y_{SE} - y_{SW}\right)^2}}, \tag{36}$$

$$dist_{E,i} = \frac{\left|\left(y_{NE} - y_{SE}\right)x_i - \left(x_{NE} - x_{SE}\right)y_i + x_{NE}y_{SE} - y_{NE}x_{SE}\right|}{\sqrt{\left(x_{NE} - x_{SE}\right)^2 + \left(y_{NE} - y_{SE}\right)^2}}, \tag{37}$$

We will then assign the detection $i$ to an area $k$ for each wall (and for any wall), according to the distance of the point to the wall as defined as follows:

$$N : i \in k \quad if \quad k\left(dst\right) < dist_{N,i} \le k\left(dst + 1\right), \tag{38}$$

$$W : i \in k \quad if \quad k\left(dst\right) < dist_{W,i} \le k\left(dst + 1\right), \tag{39}$$

$$E : i \in k \quad if \quad k\left(dst\right) < dist_{E,i} \le k\left(dst + 1\right), \tag{40}$$

$$All : i \in k \quad if \quad k\left(dst\right) < \min\left(\left\{dist_{N,i}, dist_{W,i}, dist_{S,i}, dist_{E,i}\right\}\right) \le k\left(dst + 1\right), \tag{41}$$

where *dst* (*ana.zsiz*) in an arbitrary distance given by the user to define the number of areas. Therefore, for every arena, we count all the assignments for each area and present the results in three different metrics, the total number of detections (frame count, Table 11), the total number of detections divided by the frame rate (representing the amount of seconds in each area, Table 12), and the frequency in each area (the number of detections in that area divided by the total number of detections, Table 13).

Table 11. *Dist_Edges.txt* files, total number of detections.

| | Frame Count | | | | | | | | | |
|---|---|---|---|---|---|---|---|---|---|---|
| | 20mm | 40mm | 60mm | 80mm | 100mm | 120mm | 140mm | 160mm | 180mm | 200mm |
| Edge N | 205850 | 159661 | 112987 | 89949 | 77866 | 72482 | 71472 | 67567 | 67440 | 70892 |
| Edge W | 182286 | 150407 | 115214 | 86531 | 75916 | 71105 | 67611 | 66397 | 68303 | 70350 |
| Edge S | 191830 | 146005 | 100341 | 81174 | 71692 | 67630 | 67383 | 70903 | 72280 | 77001 |
| Edge E | 212245 | 152288 | 96947 | 81260 | 73388 | 68878 | 66145 | 67615 | 69283 | 73340 |
| Edge ALL | 747518 | 412348 | 182703 | 91367 | 51580 | 28616 | 11324 | 118 | 0 | 0 |

Table 12. *Dist_Edges.txt* files, total number of detections divided by the frame rate.

| | Time Count (sec) | | | | | | | | | |
|---|---|---|---|---|---|---|---|---|---|---|
| | 20mm | 40mm | 60mm | 80mm | 100mm | 120mm | 140mm | 160mm | 180mm | 200mm |
| Edge N | 8234 | 6386.44 | 4519.48 | 3597.96 | 3114.64 | 2899.28 | 2858.88 | 2702.68 | 2697.6 | 2835.68 |
| Edge W | 7291.44 | 6016.28 | 4608.56 | 3461.24 | 3036.64 | 2844.2 | 2704.44 | 2655.88 | 2732.12 | 2814 |
| Edge S | 7673.2 | 5840.2 | 4013.64 | 3246.96 | 2867.68 | 2705.2 | 2695.32 | 2836.12 | 2891.2 | 3080.04 |
| Edge E | 8489.8 | 6091.52 | 3877.88 | 3250.4 | 2935.52 | 2755.12 | 2645.8 | 2704.6 | 2771.32 | 2933.6 |
| Edge ALL | 29900.7 | 16493.9 | 7308.12 | 3654.68 | 2063.2 | 1144.64 | 452.96 | 4.72 | 0 | 0 |

Table 13. *Dist_Edges.txt* files, frequency in each area (the number of detections in that area divided by the total number of detections.

| | Frequency (Zone Detections/Total Detections) | | | | | | | | | |
|---|---|---|---|---|---|---|---|---|---|---|
| | 20mm | 40mm | 60mm | 80mm | 100mm | 120mm | 140mm | 160mm | 180mm | 200mm |
| Edge N | 0.13 | 0.10 | 0.07 | 0.06 | 0.05 | 0.05 | 0.05 | 0.04 | 0.04 | 0.05 |
| Edge W | 0.12 | 0.10 | 0.08 | 0.06 | 0.05 | 0.05 | 0.04 | 0.04 | 0.04 | 0.05 |
| Edge S | 0.13 | 0.10 | 0.07 | 0.05 | 0.05 | 0.04 | 0.04 | 0.05 | 0.05 | 0.05 |
| Edge E | 0.14 | 0.10 | 0.06 | 0.05 | 0.05 | 0.05 | 0.04 | 0.04 | 0.05 | 0.05 |
| Edge ALL | 0.49 | 0.27 | 0.12 | 0.06 | 0.03 | 0.02 | 0.01 | 0.00 | 0.00 | 0.00 |

Note: In practice, for memory reason the matrix size will be predefined, and the output can be bigger than the actual arena being filled with ceros.

The graphic outputs, *Dist_Edge_[edge].jpeg* correspond to colored representation of the frequency of the use of each zone. To this end the matrix is normalized so the maximum frequency will be represented as 1 and the minimum none zero value will be coded as a close to cero constant. The color scale is linear and from low to high frequency, the color codes used will go from blue to green to yellow to red as shown in Figure 18. In the graphic outputs, the zones will be superimposed to the images of the arenas, and the zone size is represented in a true scale.

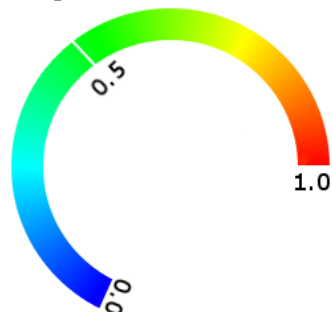

Figure 19. Color linear scale for representing the normalized of the frequency of use of a zone in the heat maps.

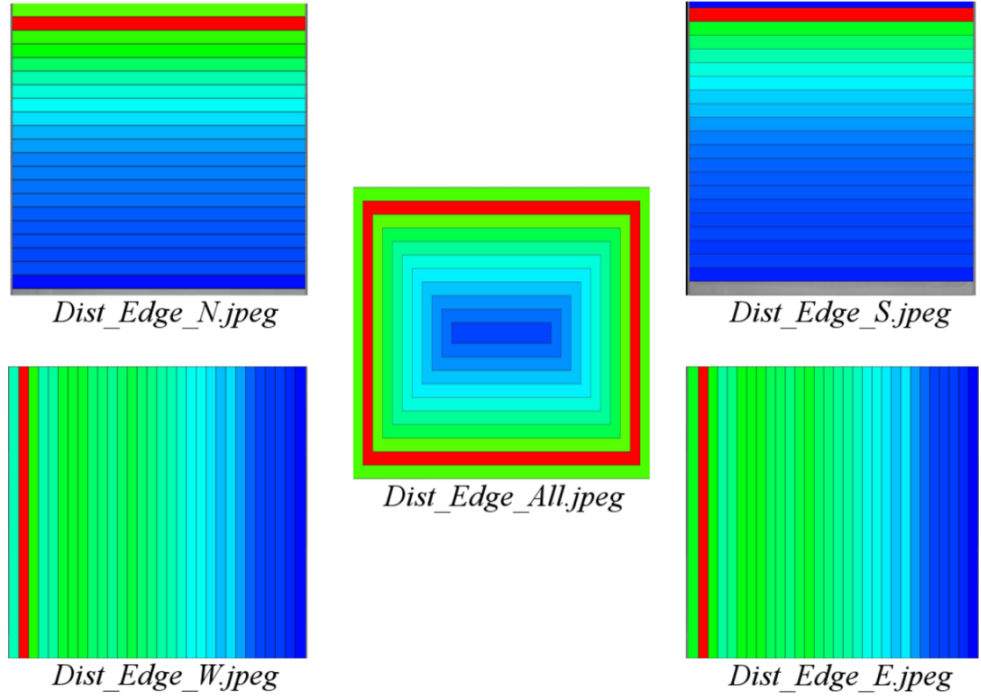

Figure 20. *Dist_Edge_[edge].jpeg*. Representation of the zone use as heat maps according to de distance to the edges.

### Exploration (*Exploration.txt, Exploration.jpeg*)

The arena is divided in regular square non overlapping zones of a size selected by user. And the use of each zone is computed for each arena in the same way as the edge zones. Therefore a matrix of zones will be created, so the zone k corresponding to the coordinates of the zone matrix will be assigned to a detected position $(x_i, y_i)$ corresponding to the time $t_i$ according to the following formula:

$$i \in k\left(k_x, k_y\right) \begin{cases} floor\left(\dfrac{\dfrac{\left|\left(y_{NE} - y_{NW}\right)x_i - \left(x_{NE} - x_{NW}\right)y_i + x_{NE}y_{NW} - y_{NE}x_{NW}\right|}{\sqrt{\left(x_{NE} - x_{NW}\right)^2 + \left(y_{NE} - y_{NW}\right)^2}}}{dst}\right) \\ floor\left(\dfrac{\dfrac{\left|\left(y_{SW} - y_{NW}\right)x_i - \left(x_{SW} - x_{NW}\right)y_i + x_{SW}y_{NW} - y_{SW}x_{NW}\right|}{\sqrt{\left(x_{SW} - x_{NW}\right)^2 + \left(y_{SW} - y_{NW}\right)^2}}}{dst}\right) \end{cases}, \qquad (42)$$

where *dst* (*ana.zsiz*) represents the size of each square of the arena, selected by the user.

For every arena, we count all the assignments for each area and present the results in three different metrics, the total number of detections (frame count, Table 14), the total number of detections divided by the frame rate (representing the amount of seconds in each area, Table 15), and the frequency in each area (the number of detections in that area divided by the total number of detections, Table 16).

Table 14. *Exploration.txt* files, total number of detections.

| | Frame Count | | | | | | | | | |
|---|---|---|---|---|---|---|---|---|---|---|
| | 20mm | 40mm | 60mm | 80mm | 100mm | 120mm | 140mm | 160mm | 180mm | 200mm |
| 20mm | 10865 | 16504 | 15511 | 13738 | 13299 | 13319 | 12400 | 12151 | 13146 | 14016 |
| 40mm | 17455 | 14569 | 11358 | 10022 | 8670 | 7977 | 7707 | 7670 | 7777 | 8864 |
| 60mm | 16034 | 11226 | 12950 | 5912 | 5334 | 4370 | 4135 | 4164 | 4206 | 4057 |
| 80mm | 15243 | 9955 | 6676 | 4223 | 3629 | 3317 | 2928 | 2627 | 2786 | 2918 |
| 100mm | 13531 | 8535 | 5285 | 3121 | 2752 | 2751 | 2508 | 2292 | 2479 | 2507 |
| 120mm | 12313 | 8443 | 5249 | 2789 | 2708 | 2641 | 2274 | 2261 | 2081 | 2176 |
| 140mm | 11653 | 8242 | 4251 | 3125 | 2214 | 2292 | 2496 | 2268 | 1996 | 2177 |
| 160mm | 10503 | 8381 | 4527 | 3424 | 2234 | 2196 | 2301 | 2263 | 1884 | 1912 |
| 180mm | 10561 | 7757 | 4627 | 3391 | 2577 | 2242 | 2234 | 2158 | 2084 | 2039 |
| 200mm | 11596 | 7637 | 5031 | 3309 | 2963 | 2838 | 2228 | 2140 | 2383 | 2299 |

Table 15. *Exploration.txt* files, total number of detections divided by the frame rate.

| | Time Count (sec) | | | | | | | | | |
|---|---|---|---|---|---|---|---|---|---|---|
| | 20mm | 40mm | 60mm | 80mm | 100mm | 120mm | 140mm | 160mm | 180mm | 200mm |
| 20mm | 434.6 | 660.2 | 620.4 | 549.5 | 532.0 | 532.8 | 496.0 | 486.0 | 525.8 | 560.6 |
| 40mm | 698.2 | 582.8 | 454.3 | 400.9 | 346.8 | 319.1 | 308.3 | 306.8 | 311.1 | 354.6 |
| 60mm | 641.4 | 449.0 | 518.0 | 236.5 | 213.4 | 174.8 | 165.4 | 166.6 | 168.2 | 162.3 |
| 80mm | 609.7 | 398.2 | 267.0 | 168.9 | 145.2 | 132.7 | 117.1 | 105.1 | 111.4 | 116.7 |
| 100mm | 541.2 | 341.4 | 211.4 | 124.8 | 110.1 | 110.0 | 100.3 | 91.7 | 99.2 | 100.3 |
| 120mm | 492.5 | 337.7 | 210.0 | 111.6 | 108.3 | 105.6 | 91.0 | 90.4 | 83.2 | 87.0 |
| 140mm | 466.1 | 329.7 | 170.0 | 125.0 | 88.6 | 91.7 | 99.8 | 90.7 | 79.8 | 87.1 |
| 160mm | 420.1 | 335.2 | 181.1 | 137.0 | 89.4 | 87.8 | 92.0 | 90.5 | 75.4 | 76.5 |
| 180mm | 422.4 | 310.3 | 185.1 | 135.6 | 103.1 | 89.7 | 89.4 | 86.3 | 83.4 | 81.6 |
| 200mm | 463.8 | 305.5 | 201.2 | 132.4 | 118.5 | 113.5 | 89.1 | 85.6 | 95.3 | 92.0 |

Table 16. *Exploration.txt* files, frequency in each area (the number of detections in that area divided by the total number of detections.

| | Frequency (Zone Detections/Total Detections) | | | | | | | | | |
|---|---|---|---|---|---|---|---|---|---|---|
| | 20mm | 40mm | 60mm | 80mm | 100mm | 120mm | 140mm | 160mm | 180mm | 200mm |
| 20mm | 0.0071 | 0.0108 | 0.0102 | 0.0090 | 0.0087 | 0.0087 | 0.0081 | 0.0080 | 0.0086 | 0.0092 |
| 40mm | 0.0114 | 0.0095 | 0.0074 | 0.0066 | 0.0057 | 0.0052 | 0.0051 | 0.0050 | 0.0051 | 0.0058 |
| 60mm | 0.0105 | 0.0074 | 0.0085 | 0.0039 | 0.0035 | 0.0029 | 0.0027 | 0.0027 | 0.0028 | 0.0027 |
| 80mm | 0.0100 | 0.0065 | 0.0044 | 0.0028 | 0.0024 | 0.0022 | 0.0019 | 0.0017 | 0.0018 | 0.0019 |
| 100mm | 0.0089 | 0.0056 | 0.0035 | 0.0020 | 0.0018 | 0.0018 | 0.0016 | 0.0015 | 0.0016 | 0.0016 |
| 120mm | 0.0081 | 0.0055 | 0.0034 | 0.0018 | 0.0018 | 0.0017 | 0.0015 | 0.0015 | 0.0014 | 0.0014 |
| 140mm | 0.0076 | 0.0054 | 0.0028 | 0.0020 | 0.0015 | 0.0015 | 0.0016 | 0.0015 | 0.0013 | 0.0014 |
| 160mm | 0.0069 | 0.0055 | 0.0030 | 0.0022 | 0.0015 | 0.0014 | 0.0015 | 0.0015 | 0.0012 | 0.0013 |
| 180mm | 0.0069 | 0.0051 | 0.0030 | 0.0022 | 0.0017 | 0.0015 | 0.0015 | 0.0014 | 0.0014 | 0.0013 |
| 200mm | 0.0076 | 0.0050 | 0.0033 | 0.0022 | 0.0019 | 0.0019 | 0.0015 | 0.0014 | 0.0016 | 0.0015 |

The graphic output, *Exploration.jpeg* represent the frequency of use of each area in the same way as explained before.

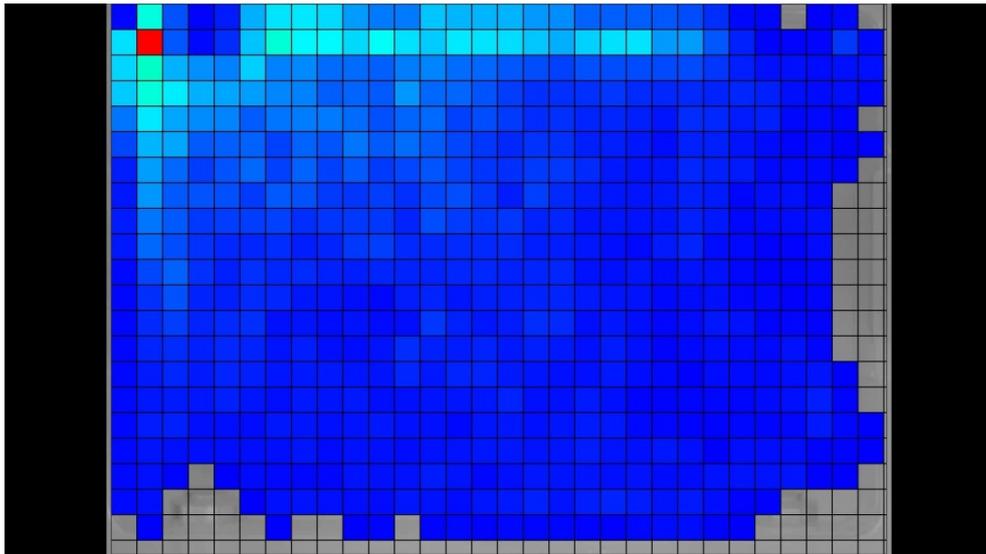

Figure 21. *Exploration.jpeg*. Representation of the zone use (exploration) as a heat map with a grid of a size selected by the user, for a single arena experimental setup with 5 fish.

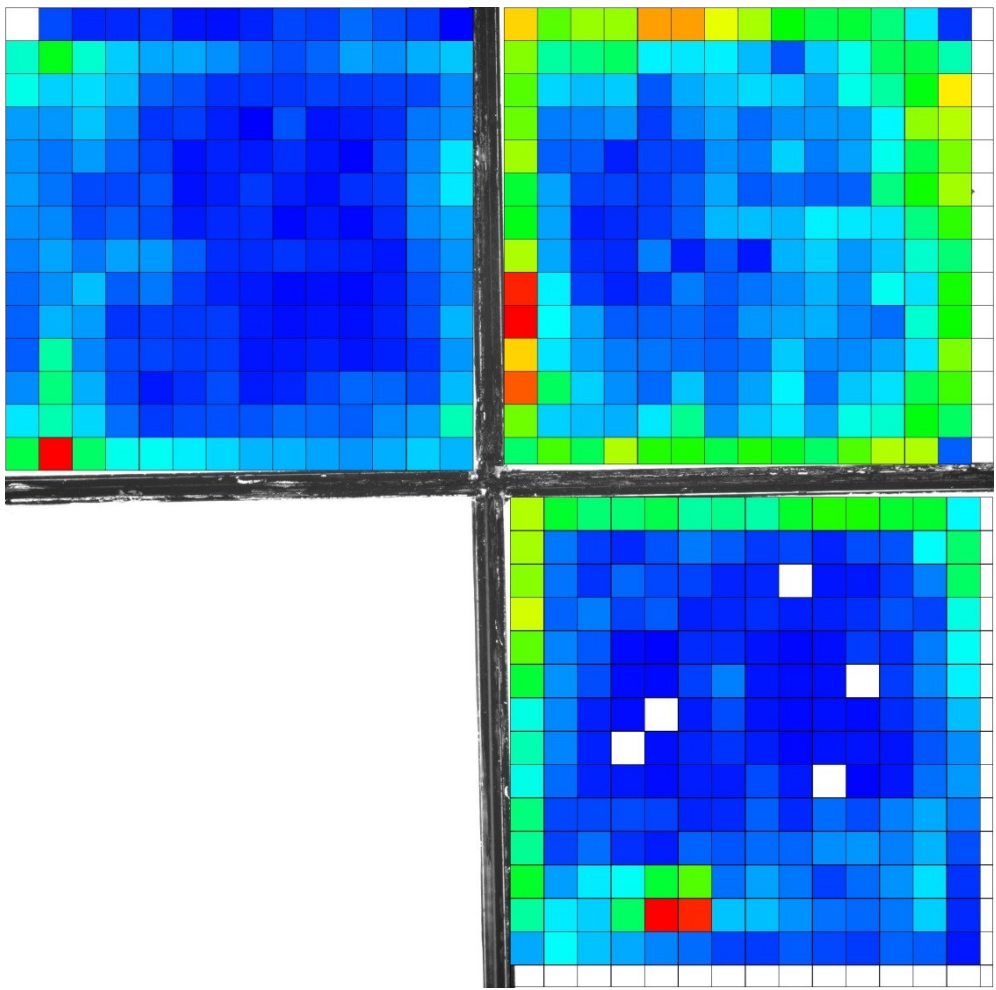

Figure 22. *Exploration.jpeg*. Representation of the zone use (exploration) as a heat map with a grid of a size selected by the user for a four arena experimental setup with three fish.

## Distance to Mean Position (*Dist_MeanPos.txt, Dist_Mean_Pos.jpeg*)

This output is similar to *Dist_Edges.txt*, representations, with the difference that the zones will be created according to the distance of a detected position *(x_i, y_i)* corresponding to the time *t_i*, to the mean of the detected positions. *(x_M, y_M)*. This is defined as follows.

$$dist_{M,i} = \sqrt{\left(x_M - x_i\right)^2 + \left(y_M - y_i\right)^2} \, , \tag{43}$$

We will assign the detection *i* to an area *k*, according to the distance of the point to the mean position defined as follows:

$$M : i \in k \quad if \quad k\left(dst\right) < dist_{M,i} \leq k\left(dst+1\right), \tag{44}$$

where *dst (ana.zsizq)* is an arbitrary distance given by the user to define the number of areas.

For every arena, we count all the assignments for each area and present the results in three different metrics, the total number of detections (frame count, Table 17), the total number of detections divided by the frame rate (representing the amount of seconds in each area, Table 18), and the frequency in each area (the number of detections in that area divided by the total number of detections, Table 19).

Table 17. *Dist_MeanPos.txt* files, total number of detections.

|  | 20mm | 40mm | 60mm | 80mm | 100mm | 120mm | 140mm | 160mm | 180mm | 200mm |
|---|---|---|---|---|---|---|---|---|---|---|
| Frame Count | 7264 | 20741 | 35441 | 59533 | 101590 | 200553 | 445979 | 371775 | 240955 | 41882 |

Table 18. *Dist_MeanPos.txt* files, total number of detections divided by the frame rate.

|  | 20mm | 40mm | 60mm | 80mm | 100mm | 120mm | 140mm | 160mm | 180mm | 200mm |
|---|---|---|---|---|---|---|---|---|---|---|
| Time Count (sec) | 290.56 | 829.64 | 1417.64 | 2381.32 | 4063.6 | 8022.12 | 17839.2 | 14871 | 9638.2 | 1675.28 |

Table 19. *Dist_MeanPos.txt* files, frequency in each area (the number of detections in that area divided by the total number of detections.

|  | 20mm | 40mm | 60mm | 80mm | 100mm | 120mm | 140mm | 160mm | 180mm | 200mm |
|---|---|---|---|---|---|---|---|---|---|---|
| Frequency (Zone Det./Total Det.) | 0.0048 | 0.0136 | 0.0232 | 0.0390 | 0.0666 | 0.1314 | 0.2923 | 0.2437 | 0.1579 | 0.0275 |

The graphic output, *Dist_Mean_Pos.jpeg* represent the frequency of use of each area in the same way as explained before.

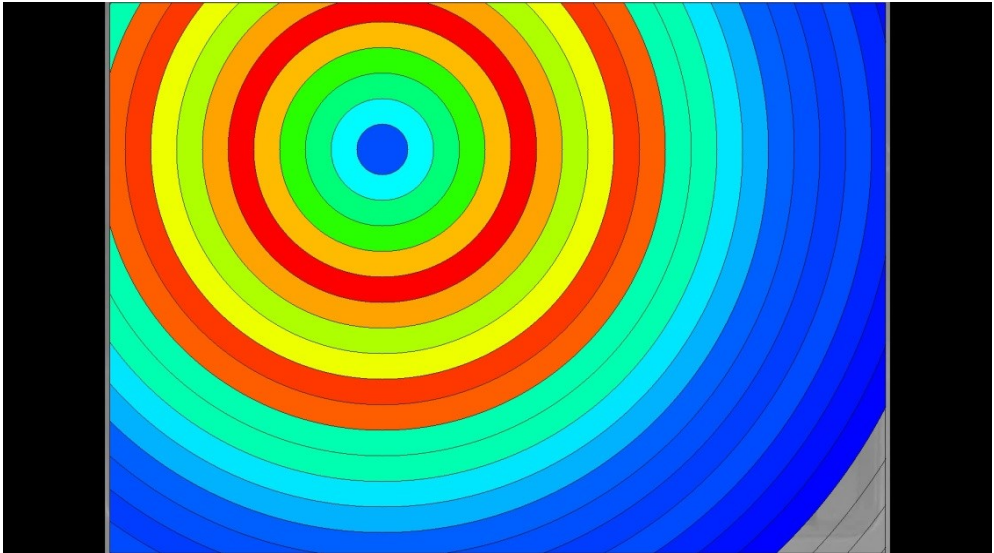

Figure 23. *Dist_Mean_Pos.jpeg*. Representation of the zone use as heat maps according to de distance to mean position of the arena, for a single arena experimental setup with 5 fish.

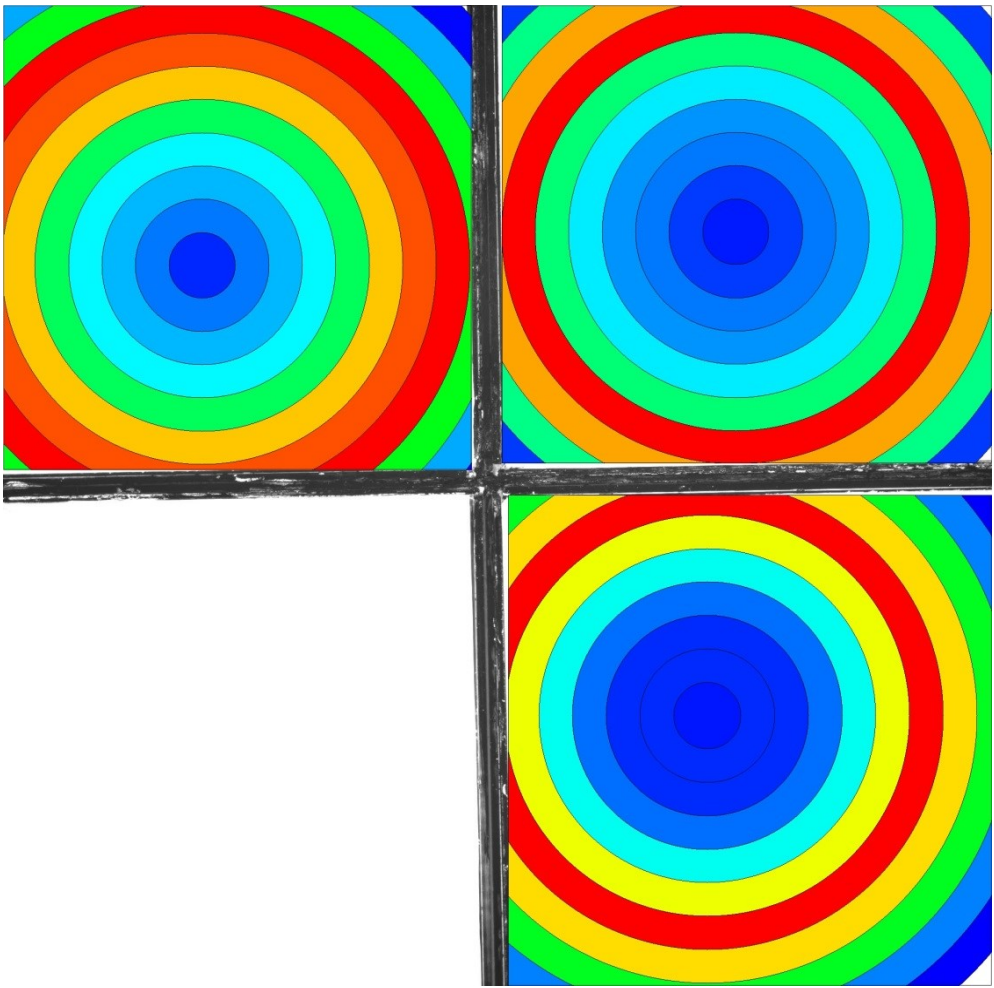

Figure 24. *Dist_Mean_Pos.jpeg*. Representation of the zone use as heat maps according to de distance to mean position of the arena, for a four arena experimental setup with three fish.

## Distance to Center Position (*Dist_CenterPos.txt, Dist_Center_Pos.jpeg*)

This output is similar to the *Dist_Edges.txt*, representations, with the difference that the zones will be created according to the distance of a detected position *(x_i, y_i)* corresponding to the time $t_i$, to the center of the arena. *(x_C, y_C)*. This is defined as follows.

$$dist_{C,i} = \sqrt{(x_C - x_i)^2 + (y_C - y_i)^2} \;, \tag{45}$$

We will assign the detection i to an area k, according to the distance of the point to center position defined as follows:

$$C : i \in k \quad if \quad k(dst) < dist_{C,i} \le k(dst+1). \tag{46}$$

where *dst* (*ana.zsizq*) is an arbitrary distance given by the user to define the number of areas.

For every arena, we count all the assignments for each area and present the results in three different metrics, the total number of detections (frame count, Table 10), the total number of detections divided by the frame rate (representing the amount of seconds in each area, Table 21), and the frequency in each area (the number of detections in that area divided by the total number of detections, Table 21).

Table 20. *Dist_CenterPos.txt* files, total number of detections.

| | 20mm | 40mm | 60mm | 80mm | 100mm | 120mm | 140mm | 160mm | 180mm | 200mm |
|---|---|---|---|---|---|---|---|---|---|---|
| Frame Count | 7226 | 20643 | 35576 | 59870 | 100260 | 199441 | 445106 | 373978 | 244069 | 39543 |

Table 21. *Dist_CenterPos.txt* files, total number of detections divided by the frame rate.

| | 20mm | 40mm | 60mm | 80mm | 100mm | 120mm | 140mm | 160mm | 180mm | 200mm |
|---|---|---|---|---|---|---|---|---|---|---|
| Time Count (sec) | 289.04 | 825.72 | 1423.04 | 2394.8 | 4010.4 | 7977.64 | 17804.2 | 14959.1 | 9762.76 | 1581.72 |

Table 22. *Dist_CenterPos.txt* files, frequency in each area (the number of detections in that area divided by the total number of detections.

| | 20mm | 40mm | 60mm | 80mm | 100mm | 120mm | 140mm | 160mm | 180mm | 200mm |
|---|---|---|---|---|---|---|---|---|---|---|
| Frequency (Zone Det./Total Det.) | 0.0047 | 0.0135 | 0.0233 | 0.0392 | 0.0657 | 0.1307 | 0.2917 | 0.2451 | 0.1600 | 0.0259 |

The graphic output, *Dist_Center_Pos.jpeg* represent the frequency of use of each area in the same way as explained before.

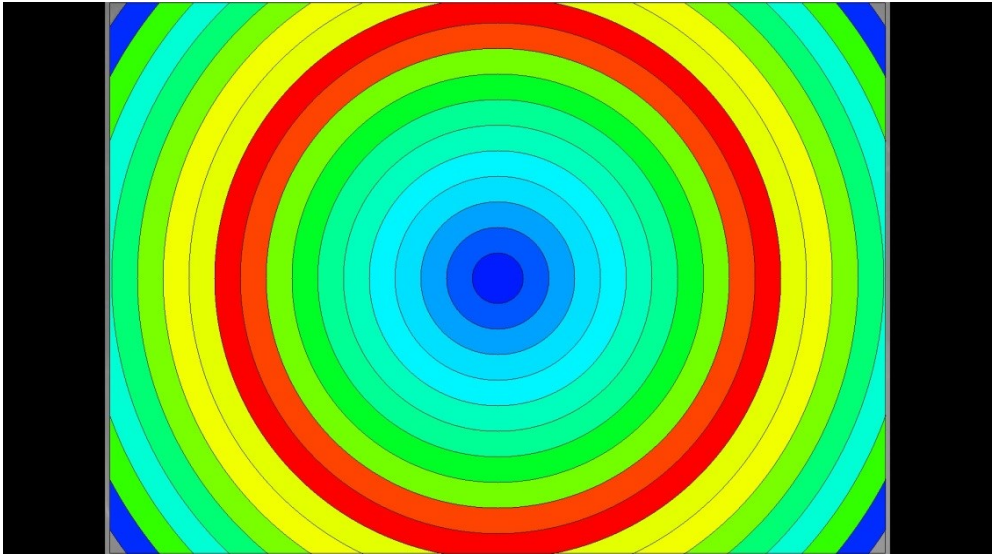

Figure 25. *Dist_Center_Pos.jpeg*. Representation of the zone use as heat maps according to de distance to central position of the arena, for a single arena experimental setup with 5 fish.

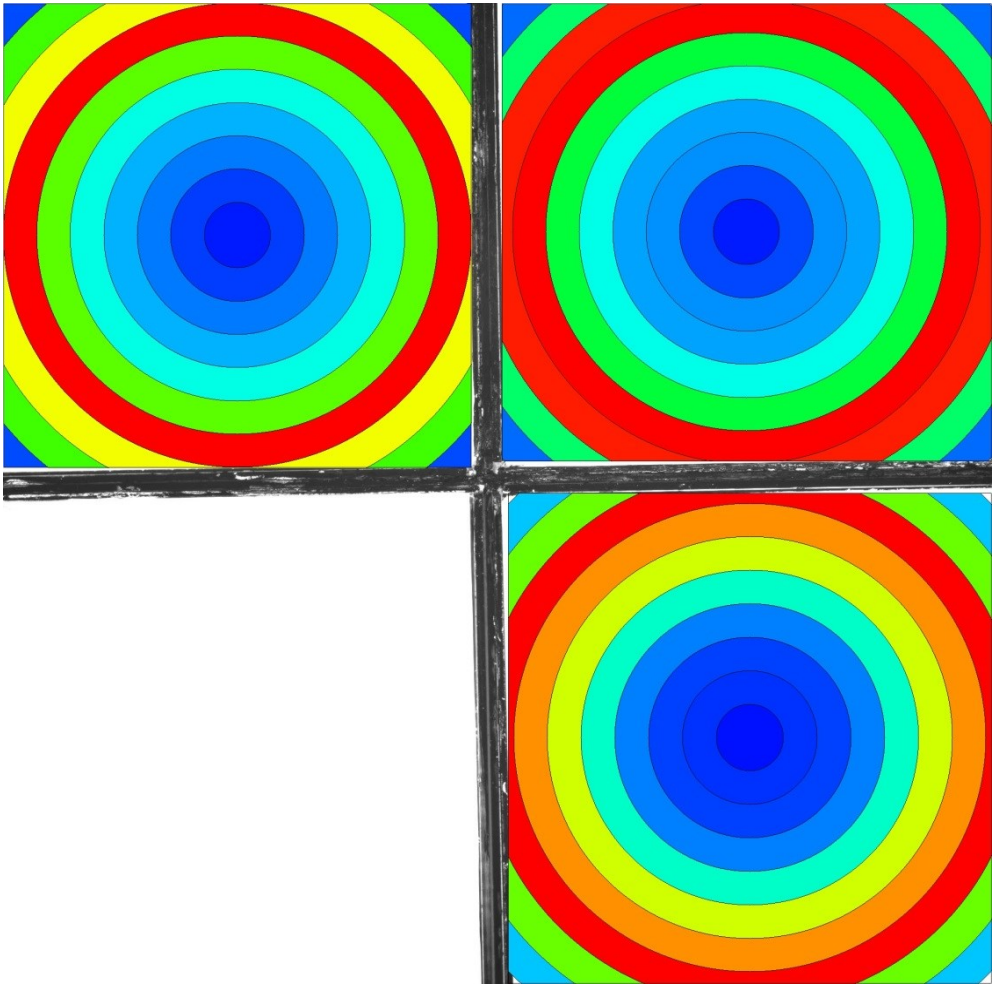

Figure 26. *Dist_Center_Pos.jpeg*. Representation of the zone use as heat maps according to de distance to central position of the arena, for a four arena experimental setup with three fish.

## Transitions (*Transitions.txt*)

Transitions, count the moments when tracked objects appears in the image (transitions labeled as 1), or disappear from it (transitions labeled as 0).

The time, track, position and other parameters from the first/last detection is saved in a file.

A transition is detected when the time between two consecutive detections exceed a value (*ana.ttim*) selected by the user (for the first and last detection the time is computed from the video start or the video end time).

Table 23. *Transitions.txt* files.

| Time (sec) | Video Seq. | Arena | Track | Pos. X (mm) | Pos. Y (mm) | Label |
|---|---|---|---|---|---|---|
| 1.92 | 0 | 1 | 1 | 30.1997 | 139.79 | 0 |
| 16.16 | 0 | 1 | 2 | 40.3914 | 142.042 | 1 |
| 25.56 | 0 | 1 | 2 | 264.587 | 142.659 | 0 |
| 36.28 | 0 | 1 | 3 | 67.851 | 142.557 | 1 |
| 43.92 | 0 | 1 | 3 | 260.846 | 141.895 | 0 |
| 59.8 | 0 | 1 | 4 | 37.3171 | 142.579 | 1 |
| 63 | 0 | 1 | 4 | 26.5074 | 141.028 | 0 |
| 75.44 | 0 | 1 | 5 | 266.598 | 142.652 | 1 |
| 78.32 | 0 | 1 | 5 | 268.247 | 141.085 | 0 |
| 86.12 | 0 | 1 | 6 | 42.6465 | 142.008 | 1 |

## Frozen Events (*FrozenEvents.txt*)

A frozen event represent the moments when the animal stands still in a predetermined position of the arena for a period of time.

The time, track number, position, length and other parameters from event are saved in a file.

A frozen event is detected when the animal has moved less than a value (*ana.fmmt*) in a certain amount of time (*ana.ftim*) selected by the user

Table 24. *FrozenEvents.txt* files.

| Time (sec) | Video Seq. | Arena | Track | Avg. Pos. X (mm) | Avg. Pos. Y (mm) | Time Length (sec) |
|---|---|---|---|---|---|---|
| 12.16 | 0 | 1 | 1 | 55.0868 | 33.9619 | 3.52 |
| 98.68 | 0 | 1 | 1 | 105.536 | 41.5467 | 3.6 |
| 37.12 | 6 | 1 | 2 | 188.977 | 74.738 | 7.32 |
| 51.92 | 6 | 1 | 2 | 195.399 | 76.9609 | 8.88 |
| 65.96 | 6 | 1 | 2 | 199.721 | 75.0917 | 4.6 |
| 79.32 | 6 | 1 | 2 | 208.338 | 75.3267 | 4 |
| 5.8 | 11 | 1 | 3 | 235.564 | 28.7672 | 8.24 |
| 21.16 | 11 | 1 | 3 | 223.176 | 27.4975 | 5 |
| 29.64 | 11 | 1 | 3 | 214.962 | 27.6328 | 5.8 |
| 39.08 | 11 | 1 | 3 | 202.831 | 27.8868 | 13.2 |

## Stats (*Stats.txt*)

The *Stats.txt* files represent a summary of all the relevant parameters of the arena (Table 25), and summarize the tracking results with global statistics (Table 26). For the population the main and standard deviation of parameters are calculated.

Table 25. *Stats.txt* files, sequence and arena information and video data.

| Parameter Name | Example of Value | Description |
|---|---|---|
| Video Resolution | [2048 x 2048] | Size in pixels of a video frame |
| Video FrameRate | 25 | Number of frames per second in the original video |
| Analysed Video Frames | 22500 | Number of processed video frames (assumed to be equal for every sequence) |
| Analysed Video Time | 900 | (Analysed Video Frames)/ (Video Framerate) |
| Arena | 1 Arena1<br>[0, 0]  [990.332, 0], [0, 991.625] [990.332, 991.625]<br>[0, 0, 1] [287.538, 0, 1] [0, 282.707, 1] [287.538, 282.707, 1] | Arena number and arena name.<br>Corners delimiting the arena in 2D space and in the 3D World, in this order:  [NW] [NE] [SW] [SE] |
| Arena Size | [1506 x 1078]<br>[1506 x 1078] | The dimensions of the arena in pixels and in real scale units. |
| Arena Center | [141.354, 143.769] | Position in mm of the center of the arena |
| Mean Position | [144.571, 4.32984] | Position in mm of the mean of all detections |

Table 26. *Stats.txt* files, tracking statistics.

| Parameter Name | Example of Value | Description |
|---|---|---|
| Av. Speed | 44.7585 | Average of *Instant_Speed* |
| Av. Accel | 68.3577 | Average of *Instant_Accel* |
| Mobility Rate | 0.999677 | Rate of *Instant_Speed* above a certain value (*ana.mobs*) |
| Visible Frames | 22111 | Number of detections in the arena |
| Visible Time | 884.471 | (Visible Frames)/( Video Framerate) |
| Invisible Frames | 372 | (Analyzed Video Frames) - ( Visible Frames) |
| Invisible Time | 14.91 | (Visible frames)/( Video Framerate) |
| First Visible Frame | 1 | First detection frame number |
| Last Visible Frame | 22481 | Last detection frame number |
| Visibility Rate | 0.983415 | (Visible Frames)/( Visible Frames+ Visible Frames) |
| Invisibility Rate | 0.0165853 | (1)-(Visibility Rate) |
| Explored Areas | 194 | Number of areas with assignments from *Exploration* |
| Number of Areas | 201 | Size of the area matrix from *Exploration,* if normalization is selected, the image areas beyond the maximum or minimum position of the animal will not be computed. |
| Exploration Rate | 0.967548 | (Number of Areas)/(Number of Explored Areas (from *Exploration*)) |
| Total Distance | 164834 | Total Swimming distance in mm |
| Transitions to White | 0 | Number of times where the animals appears in the image (from *Transitions*) |
| Transitions to Black | 0 | Number of times when the animals disappears in the image (from *Transitions*) |
| Number of Frozen Events | 1 | Count of the Frozen events from *FrozenEvens* |
| Total Time Frozen | 6.63131 | Total time in frozen state from *FrozenEvens* |
| Average Time Frozen | 1.11772 | Average time of a frozen state from *FrozenEvens* |

# Advanced parameters

## 1.  Graphic parameters (ColorIni.txt)

These parameters have only cosmetic effects, and determine different graphical parameters of the application. When executing the software (or when "New Project" option is selected), it loads as default parameters the values located in the *ColorIni.txt* from the application folder. The graphic parameters are separated in two groups and each one has an unique code for identification. Therefore, Table 27 show general parameters related to font sizes and line width.  And Table 28 defines color properties.

The width and size values are scaled in according to the image or window resolution, and the color values represent the RGB color values in inverse order (BGR), in accordance to the OpenCv representation. The graphic parameters are detailed as follows:

Table 27. *ColorIni.txt* file, general parameters.

| | GENERAL_PARAMETERS | |
|---|---|---|
| Code | Default Value | Function |
| traj.long | 25 | Length of the trajectory displayed in the tracking images. |
| roiL.widt | 2 | Line width of the rectangles showing the arena ROI. |
| traL.widt | 1 | Line width of the trajectories. |
| staL.widt | 1 | Line width of zone areas in the graphical outputs. |
| labl.size | 2 | Label size, for the tracking images |
| font.size | 2 | Font size. |

Table 28. *ColorIni.txt* file, color parameters.

| | COLOR_PARAMETERS | |
|---|---|---|
| Code | Default Value | Function |
| rea.bgnd | 255 255 255 | Color for the background of the virtual arena, for the population graphical outputs. |
| staL.colr | 0 0 0 | Color for the zone areas in the graphical outputs. |
| roiU.colr | 255 0 0 | Color for the non-selected arena ROIs |
| roiS.colr | 0 255 0 | Color for the selected arena ROI. |
| roNU.colr | 255 0 0 | Color for the non-selected arena names. |
| roNS.colr | 0 255 0 | Color for the selected arena name. |
| roiM.colr | 0 0 255 | Color for the tracking areas |

## 2. Configuration parameters (Configuration.txt)

The configuration parameters are stored in the *Configuration.txt* files generated for each project. When executing the software (or when "New Project" option is selected), it loads as default parameters the values located in the *ConfigurationIni.txt* from the application folder. The configuration parameters are separated in groups and each one has an unique code for identification, these groups are detailed in Tables 29-59. The configuration parameters are detailed as follows:

Table 29. *Configuration.txt* files, calibration parameters.

| | GENERAL PARAMETERS | | |
|---|---|---|---|
| Code | Default Value | Function | Special Values |
| exe.thre | 16 | Number of execution threads. Maximum number of parallel processes using for processing by the software | 1/ 0 = Non-Parallel Implementation |

Table 30. *Configuration.txt* files, calibration parameters.

| CALIBRATION_PARAMETERS | | | |
|---|---|---|---|
| Code | Default Value | Function | Special Values |
| cal.size | 20 | Size of each square of the checkerboard pattern used for calibration. (Available on the Interface) | |
| cal.cols | 10 | Calibration pattern cols (internally 1 is subtracted to this value to this). (Available on the Interface) | |
| cal.rows | 8 | Calibration pattern rows (internally 1 is subtracted to this value to this). (Available on the Interface) | |
| cal.dist | 1 | Calibration distortion model. (Available on the Interface) | 0=Rad3<br>1=Rad3+Tangent2<br>2=Rad3+Tangent2<br>3=Rad3+Tangent2+Prism4 |

Table 31. *Configuration.txt* files, arena definition parameters.

| ARENA_DEFINITION_PARAMETERS | | | |
|---|---|---|---|
| Code | Default Value | Function | Special Values |
| roi.mode | 0 | Mode of arena definition selection. (Available on the Interface). | 0=Automatic<br>1= Manual |
| roi.thre | 150 | Threshold value for selecting the tracking area (0:255). (Available on the Interface). | |
| roi.poly | 1 | The tracking area is approximated by a polygonal shape, this parameter specifies the approximation accuracy. This is the maximum distance between the original curve and its approximation. Only in automatic selection. (Available on the Interface). | |
| roi.elms | 7 | Size of the morphological element applied to in the closing operation of the tracking area. (Available on the Interface). | |
| roi.dilt | 1 | Number of times the erosion operation is applied. (Available on the Interface). | |
| roi.erot | 4 | Number of times the erosion operation is applied. (Available on the Interface). | |
| roi.mins | 100000 | Minimum area in pixels for creating an arena. Only in automatic selection. (Available on the Interface). | |
| roi.fite | 0 | Fit ellipse. Tries to convert the tracking area to the minimum enclosing circle of the selected pixels. Only for the manual selection. (Available on the Interface). | 0=Disabled<br>1=Enabled |
| roi.redr | 1 | If previous parameter is enabled, specifies the reduction in pixels of the radio of the minimum enclosing circle (Available on the Interface). | |

Table 32. *Configuration.txt* files, background parameters.

| BACKGROUND_PARAMETERS | | | |
|---|---|---|---|
| Code | Default Value | Function | Special Values |
| bgs.mode | 0 | Mode (Available on the Interface) | 0=Disabled<br>1=Enabled |
| bgs.nums | 500 | Sets the number of last frames that affect the background model | |
| bgs.thre | 25 | Threshold on the squared *Mahalanobis* distance between the pixel and the model to decide whether a pixel is well described by the background model. This parameter does not affect the background update. | |
| bgs.shad | 0 | Model with Shadows | 0=Disabled<br>1=Enabled |
| bgs.numg | 5 | Number of Gaussian functions used in the model. | |
| bgs.ratb | 0.99 | Background ratio (0:1) | |
| bgs.lstp | 1.E-06 | Learning rate of the background model (0:1) | <0=Automatic<br>0=Not update<br>1=Update Always |

Table 33. *Configuration.txt* files, preprocessing parameters.

| Code | Default Value | Function | Special Values |
|------|---------------|----------|----------------|
| | | PREPROCESSING_PARAMETERS | |
| pre.gfil | 5 | Gaussian filter preprocessing applied for every frame. The value is used as the size of the Gaussian filter | 0=Disabled |
| pre.norm | 0 | Enables or disables an image normalization preprocessing | 0=Disabled<br>1=Enabled |

Table 34. *Configuration.txt* files, detection parameters.

| Code | Default Value | Function | Special Values |
|------|---------------|----------|----------------|
| | | DETECTION_PARAMETERS | |
| det.type | 0 | Type of detection (not used and reserved for modding) | |
| det.thre | 90 | Threshold value for the segmentation (0:255). (Available on the Interface) | 0=Otsu value |

Table 35. *Configuration.txt* files, detection opening/closing parameters.

| Code | Default Value | Function | Special Values |
|------|---------------|----------|----------------|
| | | DETECTION_OPENING_CLOSING | |
| det.opcl | 0 | Type of morphological operation performed. | 0 = Opening (Erosion+Dilation)<br>1 = Closing (Dilation+Erosion) |
| det.elms | 3 | The size of the structuring element for the opening/closing operation | 0 = Disabled |
| det.dilt | 2 | Iterations of the dilation for the opening/closing operation | 0 = Disabled |
| det.erot | 2 | Iterations of the erosion for the opening/closing operation | 0 = Disabled |

Table 36. *Configuration.txt* files, detection dilation/erosion parameters.

| Code | Default Value | Function | Special Values |
|------|---------------|----------|----------------|
| | | DETECTION_DILATION_EROSION | |
| det.erdi | 0 | Type of morphological operation performed. | 0 = Dilation<br>1 = Erosion |
| det.elss | 0 | Size of the morphological element applied to in the erosion operation of the detected objects after closing. | 0=Disabled |
| det.ertt | 0 | Number of times the erosion operation is applied. | 0=Disabled |

Table 37. *Configuration.txt* files, detection filter parameters.

| Code | Default Value | Function | Special Values |
|------|---------------|----------|----------------|
| | | DETECTION_FILTER | |
| det.filt | 1 | Use filtering on detected bodies. (Available on the Interface). | 0=Disabled<br>1=Enabled |
| det.maxs | 1500 | Maximum size in pixels of the detected bodies (Available on the Interface). | |
| det.mins | 150 | Minimum size in pixels of the detected bodies (Available on the Interface). | |
| det.maxr | 0 | Maximum size in pixels of the minimum enclosing circle of the detected bodies. (Available on the Interface). | 0=Disabled |
| det.minr | 0 | Minimum size in pixels of the minimum enclosing circle of the detected bodies. (Available on the Interface). | 0=Disabled |
| det.mash | 0 | Maximum rate between the major and the minor radius of the minimum ellipse fitting the detected bodies. | 0=Disabled |
| det.mish | 0 | Minimum rate between the major and the minor radius of the minimum ellipse fitting the detected bodies. | 0=Disabled |
| det.minf | 0 | Minimum fill rate (ratio between the area of the minimum ellipse fitting the body and the actual number of pixels detected in the body). | 0=Disabled |

Table 38. *Configuration.txt* files, Kalman filter type.

| | | KALMAN_FILTER_TYPE | |
|---|---|---|---|
| Code | Default Value | Function | Special Values |
| kal.mode | 2 | Kalman filter algorithm (not used and reserved for modding). | |

Table 39. *Configuration.txt* files, Kalman filter parameters.

| | | KALMAN_FILTER_PARAMS | |
|---|---|---|---|
| Code | Default Value | Function | Special Values |
| kal.time | 0.25 | Time increment magnitude of the Kalman filter. | |
| kal.pron | 1.E-01 | Estimated process noise magnitude (determines expected changes in acceleration) of the Kalman filter. | |
| kal.mean | 1.E-05 | Estimated measurement noise magnitude of the Kalman filter (determines how much the Kalman filter will follow the detected positions). | |
| kal.errc | 1.E-01 | Magnitude to initialize the a posteriori error covariance matrix of the Kalman filter. | |

Table 40. *Configuration.txt* files, Kalman filter acceptance parameters.

| | | KALMAN_FILTER_ACEPTANCE | |
|---|---|---|---|
| Code | Default Value | Function | Special Values |
| kal.disf | 50 | Frame distance condition, defines the maximum allowed pixels the animal can displace between consecutive frames. (Available on the Interface). | |
| kal.sich | 0.4 | Size change condition. | kal.sich |

Table 41. *Configuration.txt* files, Kalman filter first delete condition parameters.

| | | KALMAN_FILTER_DELETE1 | |
|---|---|---|---|
| Code | Default Value | Function | Special Values |
| kal.dund | 1 | Maximum number of consecutive frames, a track has not been assigned to any detection, before mark this track as inactive. | |

Table 42. *Configuration.txt* files, Kalman filter second delete condition parameters.

| | | KALMAN_FILTER_DELETE2 | |
|---|---|---|---|
| Code | Default Value | Function | Special Values |
| kal.dage | 10 | Minimum size of a track to be used by the algorithm, tracks smaller that this size will be deleted. | |
| kal.dmax | 8 | Minimum number of detections for a track to be considered valid. | |

Table 43. *Configuration.txt* files, Kalman filter number of tracks.

| | | KALMAN_FILTER_NUMBEROFTRACKS | |
|---|---|---|---|
| Code | Default Value | Function | Special Values |
| kal.ntra | 1 | Defines the number of animals per arena. (Available on the Interface). | |

Table 44. *Configuration.txt* files, identity algorithm.

| | | KALMAN_MULTITRACKING_IDENTITY_ALGORITHM | |
|---|---|---|---|
| Code | Default Value | Function | Special Values |
| kal.idal | 0 | Selects track identification algorithm. (Available on the Interface). | 0=Not Id<br>1=Hist Only<br>2=Hist + TCM |
| kal.fdis | 0 | Frame distance condition, defines the maximum allowed pixels the animal can displace between consecutive frames, to consider two tracks compatibles. | 0=Calculated automatically |

Table 45. *Configuration.txt* files, identity algorithm, comparison parameters.

| KALMAN_MULTITRACKING_COMPARISON | | | |
|---|---|---|---|
| Code | Default Value | Function | Special Values |
| kal.cmsc | 0.2 | Minimum size change between to samples of a track to compute the identity matrix | |
| kal.corH | 1 | Use Histogram Correlation Distributions | 0=Disabled 1=Enabled |
| kal.cmhc | 0.7 | Minimum histogram correlation between to samples of a track to compute the identity matrix (For CMT algorithm) | |
| kal.corN | 100 | Number of classes in the Correlation Distributions | |
| kal.shaC | 10 | Maximum Shape alignment error (For Shape algorithms) | |
| kal.shaN | 500 | Number of classes in the Shape Distributions (used for optimization only) | |

Table 46. *Configuration.txt* files, identity algorithm, evaluation parameters.

| KALMAN_MULTITRACKING_EVALUATION | | | |
|---|---|---|---|
| Code | Default Value | Function | Special Values |
| kal.mcmp | 500 | Maximum number of comparisons, tries to limit the number of sample comparisons used for two tracks | |
| kal.mAvg | 1 | Use Avg. Use a mean metric instead of a max similarity value | 0=Disabled 1=Enabled |
| kal.gStd | 0.05 | Std. Dev of the Gaussian function ruling the decay of the weight in the Histogram Correlation Distribution | |

Table 47. *Configuration.txt* files, identity algorithm, selection parameters.

| KALMAN_MULTITRACKING_SELECTION | | | |
|---|---|---|---|
| Code | Default Value | Function | Special Values |
| kal.rGrp | 2 | Restriction to groups, Allow the first group/all groups/all tracks to be matched | 0=All tracks 1=All groups 2=First group |
| kal.hOrd | 1 | Use High order Correlation. Takes in account the similarity with all fragments, without reducing the speed. | |

Table 48. *Configuration.txt* files, identity algorithm, feature parameters.

| KALMAN_MULTITRACKING_FEATURE_PARAMETERS | | | |
|---|---|---|---|
| Code | Default Value | Function | Special Values |
| kal.hiss | 20 | Size (number of color clusters) used by the histograms. (Available on the Interface). | |
| kal.tcmr | 25 | Max distance from the body center used by the TCM maps. (Available on the Interface). | |
| kal.tcmd | 0 | Selects the data type the TCM maps will use. (Available on the Interface). | 0=8 Bits 1=16 Bits 2=32 Bits |
| kal.hist | 500 | History, maximum number of samples used by every track in the algorithm. (Available on the Interface). | |

Table 49. *Configuration.txt* files, tracking collision parameters.

| KALMAN_MULTITRACKING_COLLISION_PARAMETERS | | | |
|---|---|---|---|
| Code | Default Value | Function | Special Values |
| kal.advr | 0.8 | Distance modifier, will define the minimum advantage for two tracks with the same closest detection as $kal.disf \times kal.advr$ for detecting a collision. | |
| kal.advm | 10 | Minimum advantage value in pixels for detecting a collision between two tracks. | |
| kal.cnft | 20 | Number of frames conflicted tracks is kept in memory. A track is marked as conflicted if it collides with another track. The fusion algorithm will use these tracks to try to solve some collisions. | |

Table 50. *Configuration.txt* files, tracking fusion parameters.

| KALMAN_MULTITRACKING_FUSSION_PARAMETERS | | | |
|---|---|---|---|
| Code | Default Value | Function | Special Values |
| kal.tfmi | 5 | Minimum track age to be fused. | |
| kal.tfma | 10 | Maximum track age to be fused. | |
| kal.tdma | 10 | Maximum temporal distance between the active track and the suitable track for fusion. | |
| kal.acor | 0.6 | Minimum mean histogram correlation value between the active track and the suitable track for fusion. | |
| kal.bcor | 0.5 | Minimum best histogram correlation value between the active track and the suitable track for fusion. | |

Table 51. *Configuration.txt* files, minimum track size.

| KALMAN_MULTITRACKING_TRACK_PARAMETERS | | | |
|---|---|---|---|
| Code | Default Value | Function | Special Values |
| kal.mins | 50 | Minimum track size for a track to be marked as a long track, used in the track identification algorithm. | |

Table 52. *Configuration.txt* files, identity algorithm correlation parameters.

| KALMAN_MULTITRACKING_CORRELATION_PARAMETERS | | | |
|---|---|---|---|
| Code | Default Value | Function | Special Values |
| kal.idgb | 0 | Minimum correlation value to be accepted by the Hungarian algorithm in the track identification algorithm. The algorithm will still be able to identify the tracks if only one of the assignments has a smaller correlation value. (Available on the Interface). | |
| kal.idla | 0 | Minimum average correlation value to be accepted to identify one single long track in the track identification algorithm. (Available on the Interface). | |
| kal.idlb | 0 | Minimum best correlation value to be accepted to identify one single long track in the track identification algorithm. (Available on the Interface). | |
| kal.idsa | 0 | Minimum average correlation value to be accepted to identify one single short track in the track identification algorithm. (Available on the Interface). | |
| kal.idsb | 0 | Minimum best correlation value to be accepted to identify one single short track in the track identification algorithm. (Available on the Interface). | |

Table 53. *Configuration.txt* files, identity algorithm, maximum number of tracks.

| KALMAN_MULTITRACKING_OTHER_PARAMETERS | | | |
|---|---|---|---|
| Code | Default Value | Function | Special Values |
| kal.idff | 500 | Maximum number of tracks, kept in memory for the track identification algorithm. When the number of tracks reaches this number, the algorithm will try to identify the tracks to release memory, before continuing the tracking. | |

Table 54. *Configuration.txt* files, output parameters.

| OUTPUT_PARAMETERS | | | |
|---|---|---|---|
| Code | Default Value | Function | Special Values |
| out.step | 10 | Allows changing the number of frames in the tracking processing stage to update the interface or reaching a safe processing point (check for stop signal). | |
| out.wind | 0 | Enables to show the tracking results in real time, showing a free resizable window of the current tracking for every arena. This option is useful for demonstration or debugging, but slows down significantly the processing speed. | 0=None<br>1=Tracking<br>2=Trajectory |
| out.ftxt | 1 | Modes of txt output. Changes the amount of plain text files the software generates in the results. | 0=None<br>1=Main stats<br>2=All text files |
| out.fjpg | 1 | Enables to save an image file for every arena and frame in real time showing the tracking results. This option is useful for demonstration or debugging, but slows down significantly the processing speed. Also allows enabling or disabling the spatial stats image results. | 0=None<br>1=Spatial stats only<br>2=Tracking<br>3=Trajectory<br>4=Trajectory |
| out.pnam | TestProject | Project name, it will be used as a part of the filenames of the output files. | |

Table 55. *Configuration.txt* files, data analysis arena parameters.

| DATA_ANALYSIS_ARENA | | | |
|---|---|---|---|
| Code | Default Value | Function | Special Values |
| ana.norm | 0 | Normalize arenas. If enabled, the zones will be computed according to the extremes of the detected positions instead of using the full image. | 0=Disabled<br>1=Enabled |
| ana.aror | 3 | If the arenas are not symmetrical, and do not have the same orientation. This button allows changing the orientation of the arenas, in the projected virtual arena for the entire population. | 0=Same Orientation<br>1=Horizontal Mirror<br>2=Vertical Mirror<br>3=Vertical & Horizontal Mirror |

Table 56. *Configuration.txt* files, data analysis zone parameters.

| DATA_ANALYSIS_ZONE | | | |
|---|---|---|---|
| Code | Default Value | Function | Special Values |
| ana.nzon | 30 | Max number of areas in the edge spatial results. | |
| ana.zsizq | 50 | Distance in mm of each area in the spatial results. | |

Table 57. *Configuration.txt* files, data analysis speed parameters.

| DATA_ANALYSIS_SPEED | | | |
|---|---|---|---|
| Code | Default Value | Function | Special Values |
| ana.spsa | 2 | Sampling distance (in frames) to estimate the instantaneous speeds. | |
| ana.mobs | 1 | Speed threshold to estimate the mobility rate of the animal in the stats. | |

Table 58. *Configuration.txt* files, data analysis frozen events parameters.

| DATA_ANALYSIS_FROZEN | | | |
|---|---|---|---|
| Code | Default Value | Function | Special Values |
| ana.fmmt | 5 | A frozen event is detected when the animal has moved less than this value (in mm) in a certain amount of time (ana.ftim). | |
| ana.ftim | 3 | A frozen event is detected when the animal has moved less than ana.fmmt in this amount of time (in seconds). | |

Table 59. *Configuration.txt* files, data analysis transitions parameters.

| DATA_ANALYSIS_TRANSITIONS | | | |
|---|---|---|---|
| Code | Default Value | Function | Special Values |
| ana.ttim | 7 | A transition is detected when the time between two consecutive detections exceed this value (in seconds). | |

Table 60. *Configuration.txt* files, data analysis post process parameters.

| DATA_ANALYSIS_POSTPROCESS | | | |
|---|---|---|---|
| Code | Default Value | Function | Special Values |
| ana.inte | 0 | Interpolate holes in the trajectory, using a linear interpolation algorithm. This parameter can be changed without redoing the analysis. | 0=Disabled 1=Enabled |
| ana.intf | 25 | Maximum size of a trajectory hole (in frames) where interpolation will be applied. This parameter can be changed without redoing the analysis. | |
| ana.smoo | 0 | Use a moving average to smooth trajectories. This parameter can be changed without redoing the analysis. | 0=Disabled 1=Enabled |

Table 61. *Configuration.txt* files, data analysis other parameters.

| DATA_ANALYSIS_OTHER | | | |
|---|---|---|---|
| Code | Default Value | Function | Special Values |
| ana.rvis | 0.05 | If visibility rate is smaller than this value, normalization will not be applied even if the option is selected. | |

Table 62. *Configuration.txt* files, main video parameters.

| MAIN_VIDEO_PARAMETERS | | | |
|---|---|---|---|
| Code | Default Value | Function | Special Values |
| oth.mini | 0 | Starting point in minutes (for all sequences) for analysis. | |
| oth.mend | 1.E+10 | Ending point in minutes (for all sequences) for analysis. | |
| oth.atyp | 8 | Arena type (not used and reserved for modding). | |
| oth.frat | 25 | Frame Rate (not used and reserved for modding). | |
| oth.rees | 1 | Change of image size (not used and reserved for modding). | |